\documentclass[runningheads]{llncs}
\usepackage{graphicx}
\usepackage{comment}
\usepackage{amsmath,amssymb} 
\usepackage{color}
\usepackage{multirow}
\usepackage{lipsum}
\usepackage{booktabs}

\usepackage{subcaption}
\usepackage{rotating}

\usepackage{hyperref}

\begin{document}
\pagestyle{headings}
\mainmatter
\def\ECCVSubNumber{001}  

\title{AIM 2020 Challenge on  Real Image Super-Resolution: Methods and Results} 

\titlerunning{AIM 2020 Challenge on  Real Image Super-Resolution: Methods and Results}
\author{Pengxu Wei, Hannan Lu, Radu Timofte, Liang Lin, Wangmeng Zuo, \\
Zhihong Pan, Baopu Li, Teng Xi, Yanwen Fan, Gang Zhang, Jingtuo Liu, Junyu Han, Errui Ding,
Tangxin Xie, Liang Cao, Yan Zou, Yi Shen, Jialiang Zhang, Yu Jia,
Kaihua Cheng, Chenhuan Wu,
Yue Lin, Cen Liu, Yunbo Peng,
Xueyi Zou,
Zhipeng Luo, Yuehan Yao, Zhenyu Xu,
Syed Waqas Zamir, Aditya Arora, Salman Khan, Munawar Hayat, Fahad Shahbaz Khan,
Keon-Hee Ahn, Jun-Hyuk Kim, Jun-Ho Choi, Jong-Seok Lee,
Tongtong Zhao, Shanshan Zhao,
Yoseob Han, Byung-Hoon Kim, JaeHyun Baek,
Haoning Wu, Dejia Xu,
Bo Zhou,
Wei Guan, Xiaobo Li, Chen Ye,
Hao Li, Haoyu Zhong, Yukai Shi, Zhijing Yang, Xiaojun Yang,
Haoyu Zhong,
Xin Li, Xin Jin, Yaojun Wu, Yingxue Pang, Sen Liu,
Zhi-Song Liu, Li-Wen Wang, Chu-Tak Li, Marie-Paule Cani, Wan-Chi Siu,
Yuanbo Zhou,
Rao Muhammad Umer, Christian Micheloni,
Xiaofeng Cong,
Rajat Gupta,
Keon-Hee Ahn, Jun-Hyuk Kim, Jun-Ho Choi, Jong-Seok Lee,
Feras Almasri, Thomas Vandamme, Olivier Debeir
}
\institute{}
\authorrunning{Pengxu Wei, Hannan Lu, Radu Timofte, Liang Lin, Wangmeng Zuo, et al.}

\maketitle

\newcommand\blfootnote[1]{%
\begingroup
\renewcommand\thefootnote{}\footnote{#1}%
\addtocounter{footnote}{-1}%
\endgroup
}

\begin{abstract}
This paper introduces the real image Super-Resolution (SR) challenge that was part of the Advances in Image Manipulation (AIM) workshop, held in conjunction with ECCV 2020. This challenge involves three tracks to super-resolve an input image for $\times$2, $\times$3 and $\times$4 scaling factors, respectively. The goal is to attract more attention to realistic image degradation for the SR task, which is much more complicated and challenging, and contributes to real-world image super-resolution applications.
452 participants were registered for three tracks in total, and 24 teams submitted their results. They gauge the state-of-the-art approaches for real image SR in terms of PSNR and SSIM.
\blfootnote{P. Wei, H. Lu, R. Timofte, L. Lin and W. Zuo are the challenge organizers and the other others participated in the challenge. Appendix \textcolor{red}{A} contains the authors's teams and affiliations. AIM webpage: \url{https://data.vision.ee.ethz.ch/cvl/aim20/}}
\end{abstract}

\section{Introduction}
Single image super-resolution (SR) reconstructs high-resolution (HR) images from low-resolution (LR) counterparts with image quality degradations \cite{glasner2009super}\cite{yang2010image}. Instead of imposing higher requirements on hardware devices and sensors, it could be applicable to many practical scenarios, such as video surveillance, satellite, medical imaging, \emph{etc}. As a fundamental research topic, SR has attracted a long-standing and considerable attention in computer vision community.

With the emergence of deep learning, convolutional neural network (CNN) based SR methods (\emph{e.g.}, SRCNN \cite{SRCNN}, SRGAN \cite{SRRESNET}, EDSR \cite{EDSR}, ESRGAN \cite{ESRGAN} and RCAN \cite{RCAN}) inherit the powerful capacity of deep learning and have achieved remarkable performance improvements. Nevertheless, so far, the remarkable progress of SR is mainly driven by the supervised learning of models from LR images and their HR counterparts. While the bicubic downsampling is usually adopted to simulate the LR images, the learned deep SR model performs much less effective for real-world SR applications since the image degradation in real-world is much more complicated.

To mitigate this issue, several real SR datasets have been recently built, City 100~\cite{city100} and SR-RAW~\cite{zoomlearn}. The images in City100 were captured for the printed postcards in the indoor environment , which are limited in capturing the complicated image and degradation characteristics of natural scenes. The images in SR-RAW were collected in the real world and a contextual bilateral loss was proposed to address the misalignment problem in the dataset. Besides, Cai \emph{et al.} \cite{cai2019toward} released another real image SR dataset, named RealSR, which was captured from two DSLR cameras. They proposed the LP-KPN method in a Laplacian pyramid framework. Considering the complex image degradation across different scenes and devices, a large-scale diverse real SR dataset, named DRealSR~\cite{CDC}, was released to further promote the research on real-world image SR. Images of DRealSR were captured by five different DSLR cameras and posed more challenging image degradation. In \cite{CDC}, the proposed component divide-and-conquer model (CDC) built a baseline, hourglass SR network (HGSR), in a stacked architecture, explored different reconstruction difficulties in terms of three low-level image components inspired by corner point detection, \emph{i.e}, the flat, edges and corner points, and trained the model with a mediate supervision strategy. Besides, its proposed gradient-weighted (GW) loss also drives the model to adapt learning objectives to the reconstruction difficulties of three image components and has a flexibility of the application to any SR model.

Jointly with the Advances in Image Manipulation (AIM) 2020 workshop, we organize the AIM Challenge on Real-world Image Super-Resolution. Specifically, this challenge concerns the real-world SISR, which poses two challenging issues \cite{CDC}: (1) more complex degradation against bicubic downsampling, and (2) diverse degradation processes among devices, aiming to learn a generic model to super-resolve LR images captured in practical scenarios. To achieve this goal, paired LR and HR images are captured by various DSLR cameras and provided for training. They are randomly selected from the DRealSR dataset. Images for training, validation and testing are captured in the same way with the same set of cameras.

This challenge is one of the AIM 2020 associated challenges on:
scene relighting and illumination estimation\cite{elhelou2020aim_relighting}, image extreme inpainting\cite{ntavelis2020aim_inpainting}, learned image signal processing pipeline\cite{ignatov2020aim_ISP}, rendering realistic bokeh\cite{ignatov2020aim_bokeh}, real image super-resolution\cite{wei2020aim_realSR}, efficient super-resolution\cite{zhang2020aim_efficientSR}, video temporal super-resolution\cite{son2020aim_VTSR} and video extreme super-resolution\cite{fuoli2020aim_VXSR}.

\section{AIM 2020 Challenge on Real Image Super-Resolution }
The objectives of the AIM 2020 challenge on real image super-resolution challenge are: (i) to further explore the researches on real image SR; (ii) to fully evaluate different SR approaches on different scale factors; (iii) to offer an opportunity of communications between academic and industrial participants.

\subsection{DRealSR Dataset}

DRealSR\footnote{The dataset is publicly available at \url{https://github.com/xiezw5/Component-Divide-and-
Conquer-for-Real-World-Image-Super-Resolution}} ~\cite{CDC} is a large-scale real-world image super-resolution. Only half of images in DRealSR are randomly selected for this challenge. These images are captured from five DSLR cameras (\emph{i.e.,} Canon, Sony, Nikon, Olympus and Panasonic) in natural scenes and cover indoor and outdoor scenes avoiding moving objects, \emph{e.g.}, advertising posters, plants, offices, buildings, \emph{etc}. These HR-LR image pairs are aligned. To get access to the training and validation data and submit SR results, the registration on Codalab\footnote{\url{https://competitions.codalab.org}} is required. Details of the dataset in this challenge are given in Table~\ref{tab:dataset}.


\begin{table}[t]
\centering
\caption{Details of the dataset for the challenge}
\begin{tabular}{|c|c|c|c|c|c|}
\hline
\textbf{Scale} & \textbf{Split} & \textbf{Type} & \textbf{Number} & \textbf{Size   (LR)} & \textbf{Evaluation} \\ \hline
\hline
\multirow{3}{*}{$\times$2} & Train & Cropped Patches & 19,000 & 380$\times$380 & \multirow{9}{*}{\begin{tabular}[c]{@{}c@{}}PSNR \\(on RGB channels), \\    \\ SSIM\end{tabular}} \\ \cline{2-5}
 & Validation & Aligned Images & 20 & \multirow{2}{*}{$\sim$2000$\times$3000} &  \\ \cline{2-4}
 & Test & Aligned Images & 60 &  &  \\ \cline{1-5}
\multirow{3}{*}{$\times$3} & Train & Cropped Patches & 19,000 & 272$\times$272 &  \\ \cline{2-5}
 & Validation & Aligned Images & 20 & \multirow{2}{*}{$\sim$1300$\times$2000} &  \\ \cline{2-4}
 & Test & Aligned Images & 60 &  &  \\ \cline{1-5}
\multirow{3}{*}{$\times$4} & Train & Cropped Patches & 19,000 & 192$\times$192 &  \\ \cline{2-5}
 & Validation & Aligned Images & 20 & \multirow{2}{*}{$\sim$1000$\times$1250}  &  \\ \cline{2-4}
 & Test & Aligned Images & 60 &  &  \\ \hline
\end{tabular}
\label{tab:dataset}
\end{table}

\subsection{Track and Competition}\label{sec:track}
\textbf{Tracks.} The challenge uses the newly released DRealSR dataset and has three tracks corresponding to $\times$2, $\times$3, $\times$4 upscaling factors. The aim is to obtain a network design or solution capable to produce high-quality results with the best fidelity to the reference ground truth.

\vspace{0.5em}
\noindent
\textbf{Challenge phases.}
\emph{(1) Development phase:} HR images from DRealSR have 4000$\times$6000 pixels on average. For the convenience of model training, images are cropped into patches. For $\times$2 scale factor, LR image patches are 380$\times$380; for $\times$3 scale factor, LR image patches are 272$\times$272; for $\times$4 scale factor, LR image patches are 192$\times$192.
\emph{(2) Testing phase:} In the final test phase, participants have access to LR images for three tracks, submit their SR results to Codalab evaluation server and email their codes and factsheets to the organizers. The organizers checked all the SR results and the provided codes to obtain the final results. 

\begin{table}[!ht]
    \scriptsize
    \centering
    \caption{Evaluation results in the final testing phase. ``Score'' indicates the weighted score (Equ.\ref{Eq:evaluation}), \emph{i.e.}, the evaluation metric for the challenge. For ``Ensemble'', ``model'' and ``self'' indicate the model ensemble and the self-ensemble, respectively. ``/'' indicates that those items are not provided by participants. We also provide results of ``EDSR*'' for comparison with the same challenge dataset.}
    \label{tab:evaluation}
    \begin{tabular}{lccccccc}
        \toprule
        \textbf{Team}& \textbf{PSNR}& \textbf{SSIM}& \textbf{Score} & \textbf{Ensemble} & \textbf{ExtraData} & \textbf{Loss}\\
        \midrule
        \multicolumn{7}{c}{\textbf{Track1} ($\times2$)} \\
        \hline
        Baidu  & \textbf{33.446} & \textbf{0.927} & \textbf{0.7736} & Model+Self & False & $L_1$ + SSIM \\
        CETC-CSKT & 33.314 & 0.925 & 0.7702 & Model+Self & False & $L_1$ \\
        OPPO\_CAMERA & 33.309 & 0.924 & 0.7699 & Model+Self & False & $L_1$ + SSIM + MS-SSIM \\
        AiAiR & 33.263 & 0.924 & 0.7695 & Model+Self & True & Clip $L_1$ \\
        TeamInception & 33.232 & 0.924 & 0.7690 & Model+Self & True & $L_1$ + MS-SSIM + VGG \\
        Noah\_TerminalVision & 33.289 & 0.923 & 0.7686 & Self & False & adaptive robust loss \\
        DeepBlueAI & 33.177 & 0.924 & 0.7681 & Self & False & / \\
        ALONG  & 33.098 & 0.924 & 0.7674 & Self & False &  $L_1$ + $L_2$ \\
        LISA-ULB & 32.987 & 0.923 & 0.7659 & / & False & $L_1$ + SSIM \\
        lyl & 32.937 & 0.921 & 0.7635 & / & False & $L_1$ \\
        GDUT-SL & 32.973 & 0.920 & 0.7634 & Model & False & $L_1$ \\
        MCML-Yonsei & 32.903 & 0.919 & 0.7612 & None & False & $L_1$ \\
        Kailos & 32.708 & 0.920 & 0.7601 & Self & False & $L_1$ + wavelet loss \\
        qwq & 31.640 & 0.913 & 0.7436 & None & False & $L_1$ + SSIM \\
        debut\_kele & 31.236 & 0.889 & 0.7196 & None & True & / \\
        EDSR* & 31.220  & 0.889  & 0.7194  & / & / & / \\
        RRDN\_IITKGP & 29.851 & 0.845 & 0.6696 & None & True & / \\
        \hline
        \multicolumn{7}{c}{\textbf{Track2} ($\times3$)} \\
        \hline
        Baidu & \textbf{30.950} & \textbf{0.876} & \textbf{0.7063} & Model+Self & False & $L_1$ + SSIM \\
        CETC-CSKT & 30.765 & 0.871 & 0.7005 & Model+Self & False & $L_1$ \\
        OPPO\_CAMERA & 30.537 & 0.870 & 0.6966 & Model+Self & False & $L_1$ + SSIM + MS-SSIM \\
        Noah\_TerminalVision & 30.564 & 0.866 & 0.6941 & Self & False & adaptive robust loss \\
        MCML-Yonsei & 30.477 & 0.866 & 0.6931 & Self & False & $L_1$ \\
        TeamInception & 30.418 & 0.866 & 0.6928 & Model+Self & True & $L_1$ + MS-SSIM + VGG \\
        ALONG & 30.375 & 0.866 & 0.6922 & Self & False &  $L_1$ + $L_2$ \\
        DeepBlueAI & 30.302 & 0.867 & 0.6918 & Self & False & / \\
        lyl & 30.365 & 0.864 & 0.6905 & / & False & $L_1$ \\
        Kailos & 30.130 & 0.866 & 0.6900 & Self & False & $L_1$ + wavelet loss \\
        qwq & 29.266 & 0.852 & 0.6694 & None & False & $L_1$ + SSIM \\
        EDSR* & 28.763  & 0.821  & 0.6383  & / & / & / &  \\
        anonymous & 18.190 & 0.825 & 0.5357 & / & False & / \\
        \hline
        \multicolumn{7}{c}{\textbf{Track3} ($\times4$)} \\
        \hline
        Baidu & \textbf{31.396} & \textbf{0.875} & \textbf{0.7099} & Model+Self & False & $L_1$ + SSIM \\
        ALONG & 31.237 & 0.874 & 0.7075 & Self & False & $L_1$ + $L_2$ \\
        CETC-CSKT & 31.123 & 0.874 & 0.7066 & Model+Self & False & $L_1$ \\
        SR-IM & 31.174 & 0.873 & 0.7057 & Self & False & / \\
        DeepBlueAI & 30.964 & 0.874 & 0.7044 & Self & False & / \\
        JNSR & 30.999 & 0.872 & 0.7035 & Model+Self & True & / \\
        OPPO\_CAMERA & 30.86 & 0.874 & 0.7033 & Model+Self & False & $L_1$ + SSIM + MS-SSIM \\
        Kailos & 30.866 & 0.873 & 0.7032 & Self & False & $L_1$ + wavelet loss \\
        SR\_DLu & 30.605 & 0.866 & 0.6944 & Self & False & / \\
        Noah\_TerminalVision & 30.587 & 0.866 & 0.6944 & Self & False & adaptive robust loss \\
        Webbzhou & 30.417 & 0.867 & 0.6936 & None & False & / \\
        TeamInception & 30.347 & 0.868 & 0.6935 & Model+Self & True & $L_1$ + MS-SSIM + VGG \\
        lyl & 30.319 & 0.866 & 0.6911 & / & False & $L_1$ \\
        MCML-Yonsei & 30.420 & 0.864 & 0.6906 & Self & False & $L_1$ \\
        MoonCloud & 30.283 & 0.864 & 0.6898 & Model + Self & True & / \\
        qwq & 29.588 & 0.855 & 0.6748 & None & False & $L_1$ + SSIM \\
        SrDance & 29.595 & 0.852 & 0.6729 & / & True & MAE+VGG+GAN loss \\
        MLP\_SR & 28.619 & 0.831 & 0.6457 & Self & True & GAN,TV,$L_1$,SSIM,MS-SSIM,Cycle \\
        EDSR* & 28.212  & 0.824  & 0.6356  & / & / & / &  \\
        RRDN\_IITKGP & 27.971 & 0.809 & 0.6201 & None & True & / \\
        congxiaofeng & 26.392 & 0.826 & 0.6187 & None & False & $L_1$ \\
        \bottomrule
    \end{tabular}
\end{table}

\vspace{0.5em}
\noindent
\textbf{Evaluation protocol.}
The evaluation includes the comparison of the super-resolved images with the reference ground truth images. We use the standard peak signal to noise ratio (PSNR) and, complementary, the structural similarity (SSIM) index as often employed in the literature. PSNR and SSIM implementations are found in most of the image processing toolboxes. For each dataset, we report the average results (i.e. $PSNR_{avg}$ and $SSIM_{avg}$) over all the processed images belonging to it and employ for ranking the weighted value of normalized $PSNR_{avg}$ and $SSIM_{avg}$, which is defined as follows,

\begin{align}
{PSNR_{avg}/50+(SSIM_{avg}-0.4)/0.6}.
\label{Eq:evaluation}
\end{align}

\section{Challenge Results}
There are 174, 128 and 168 registered participants for three tracks, respectively. In total, 24 teams submitted their super-resolution results; 10, 2 and 11 teams submitted results of one, two and three tracks, respectively. Among those submitted results of one track, seven teams are for $\times4$ scale factor. Details of final testing results are provided in Table \ref{tab:evaluation}. It mainly reports the final evaluation results and model training details. 

As for the evaluation metric of weighted score claimed in Sec.\ref{sec:track}, the leading entries for Track 1, 2 and 3 are all from team Baidu. For Track 1 and 2, the CETC-CSKT and the OPPO\_CAMERA team win the second and the third places, respectively. For Track 3, ALONG and CETC-CSKT win the second and the third places, respectively.
%
Among those solutions for the challenge, some interesting trends can be observed as follows.

\textbf{Network Architecture.} All the teams utilize deep neural networks for super-resolution. The architecture of the deep network will greatly affect the performance of super-resolution images.
Several teams, \emph{e.g.}, TeamInception, construct a network with the residual structure to reduce the difficulty of optimization, While OPPO\_CAMERA connected the input to the output with a trainable convolution layer. CETC-CSKT further proposed to pre-train the trainable layer in the skip branch in advance. Several teams, such as DeepBlueAI and SR-IM applied channel attention module in their network, while several others like TeamInception and Noah\_TerminalVision employ both spatial attention and channel attention on the feature level.

\textbf{Data Augmentation.}
Most solutions conduct the data augmentation by randomly flipping and rotating images by 90 degrees. The newly proposed CutBlur method was employed by ALONG and OPPO\_CAMERA and performance improvements are reported by these teams.

\textbf{Ensemble Strategy.}
Most solutions adopted self-ensemble x8. Some solutions also performed model-ensemble by fusing results from models with different training parameter, or even of different architectures.

\textbf{Platform.}
All the teams except one team using Tensorflow utilized PyTorch to conduct their experiments.

\section{Challenge Methods and Teams}
\textbf{Baidu}

\noindent
The Baidu team proposed to apply Neural Architecture Search (NAS) approach selecting variations of their previous dense residual model as well as RCAN model\cite{zhihong2020aim}. In order to accelerate the searching process, Gaussian Process based Neural Architecture Search (GP-NAS) was applied as in \cite{Li_2020_CVPR}. Specifically, given the hyper-parameters of GP-NAS, they are capable of predicting the performance of any architectures in the search space effectively. Then, the NAS process is converted to hyper-parameters estimation. By mutual information maximization, the Baidu team can efficiently sample networks. Accordingly, based on the performances of sampled networks, the posterior distribution of hyper-parameters can be gradually and efficiently updated. Based on the estimated hyper-parameters, the architecture with the best performance can be obtained.

\begin{figure}[t]
\centering
\includegraphics[width=1\linewidth]{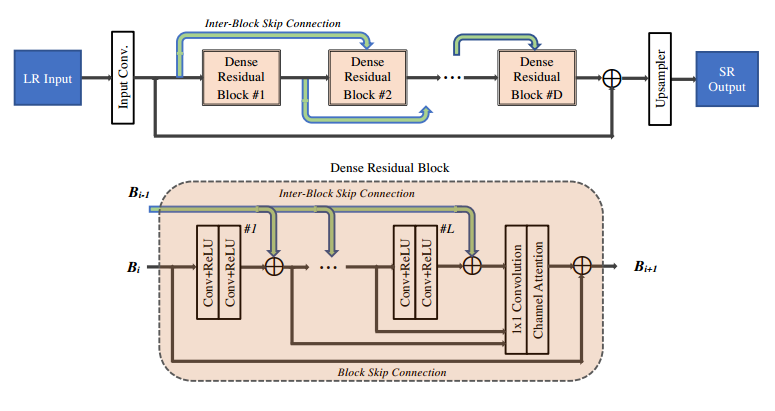}
\caption{The dense residual network architecture of the Baidu team for image Super-Resolution}
\label{fig:Baidu_DSR}
\end{figure}

The backbone model of the proposed method is a deep dense residual network originally developed for raw image demosaicking and denoising. As depicted in Fig.\ref{fig:Baidu_DSR}, in addition to the shallow feature convolution at the front and the upsampler at the end, the proposed network consists of a total depth of $D$ dense residual blocks (DRB). The input convolution layer converts the 3-channel LR input to a total of F-channel shallow features. For the middle DRB blocks, each one includes $L$ stages of double layers of convolution and the outputs of all $L$ stages are concatenated together before convoluted from $F\times L$ to F channels. An additional channel-attention layers are included at the end of each block, similar to RCAN~\cite{RCAN}. There are two types of skip connections included in each block, the block skip connection (BSC) and inter-block skip connection (IBSC). The BSC is the shortcut between input and output of block $B_i$, while IBSC includes two shortcuts from the input of block $B_{i-1}$ to the two stages inside block $B_{i}$, respectively. The various skip connections, especially IBSC, are included to combine features with a large range of receptive fields. The last block is an enhanced upsampler that transforms all F-channel LR features to the estimated 3-channel SR image. This dense residual network has three main hyper-parameters: $F$ is the number of feature channels, $D$ is the number of DRB layers and $L$ is the number of stages for each DRB. All these three hyper-parameters construct the search space for NAS.

During training, a 120$\times$120 patch is randomly cropped and augmented with flipping and transposing from each training image for each epoch. A mixed loss of $L_1$ and multi-scale structural similarity (MS-SSIM) is taken for training. For the experiment, the new model candidate search scheme using GP-NAS was implemented in PaddlePaddle~\cite{Paddle} and the final-training of searched
models were conducted using PyTorch. A multi-level ensemble scheme is proposed in testing, including self-ensemble for patches, as well as patch-ensemble and model-ensemble for full-size images. The proposed method is validated to be highly effective, generating impressive testing results on all three tracks of AIM2020 Real Image Super-resolution Challenge.

\vspace{1em}
\noindent
\textbf{CETC-CSKT}

\begin{figure}[t]
\centering
\includegraphics[scale=0.3]{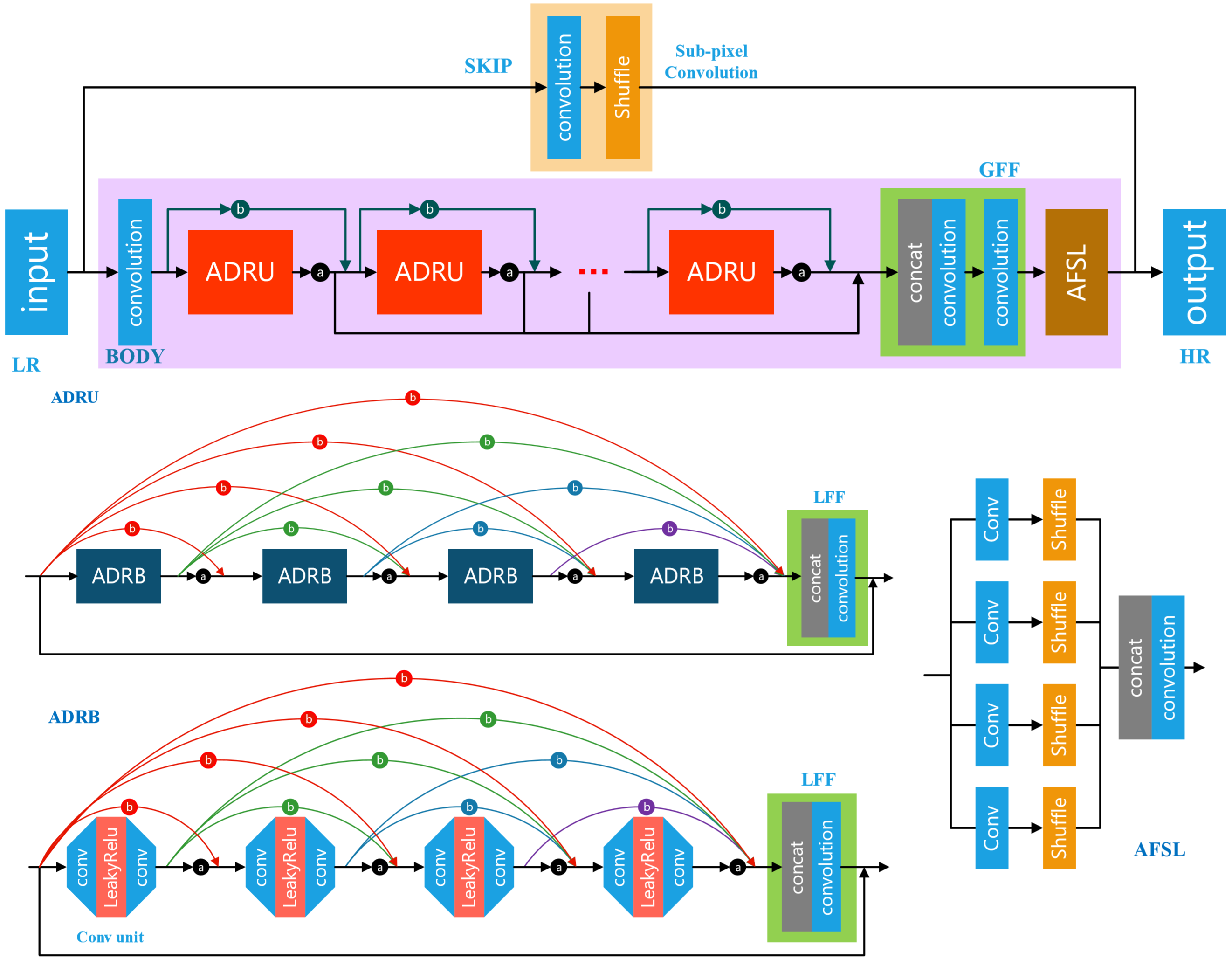}
\caption{Framework of Adaptive Dense Connection Super Resolution reconstruction (ADCSR) for the CETC-CSKT team}
\label{fig:CETC-CSKT_model}
\end{figure}

\noindent
The CETC-CSKT team proposed Adaptive Dense Connection Super Resolution reconstruction(ADCSR)\cite{xie2019adaptive}. The algorithm is divided into BODY and SKIP. The BODY part improves the utilization of convolution features through adaptive dense connection. An adaptive sub-pixel reconstruction module (AFSC) is also proposed to reconstruct the features of BODY output. By pre-training SKIP in advance, the BODY part focuses on high-frequency feature learning. for track 1 ($\times$2), spatial attention is added after each residual block. The architecture is shown in Fig.\ref{fig:CETC-CSKT_model}. Self-ensemble is used in EDSR~\cite{EDSR}. The test image is divided into {$80\times80$} pixel blocks for reconstruction. Finally, only {$60\times60$} input is used for splicing to reduce the edge difference of blocks.

The proposed ADCSR uses the first 18900 training data sets for training, and the last 100 as the test set for training. The input image block size is $80\times80$ . SKIP is trained separately, and then the entire network is trained at the same time. The initial learning rate is $1\times10^{-4}$ . When the learning rate drops to $5\times10^{-7}$, the training stops. $L_1$ loss is utilized to optimize the proposed model. The model is trained with NVIDIA RTX2080Ti * 4. Pytorch1.1.0 + Cuda10.0 + cudnn7.5.0 is selected as the deep learning environment.

\vspace{1em}
\noindent
\textbf{OPPO\_CAMERA}
\begin{figure}[t]
    \centering
    \includegraphics[width=0.9\textwidth]{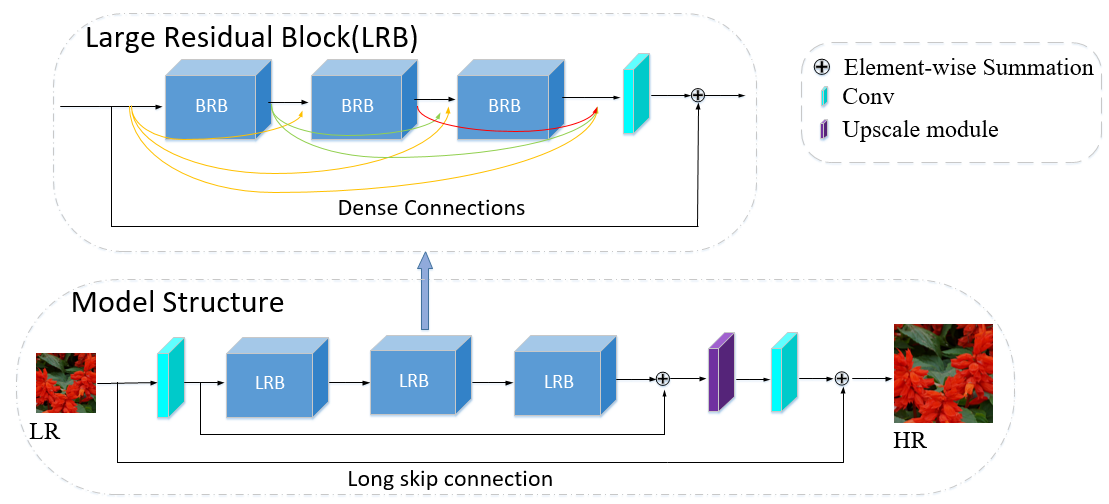}
    \caption{The detailed network architecture of the proposed network for the OPPO\_CAMERA team}
    \label{total_structure}
\end{figure}
\begin{figure}[t]
    \centering
    \includegraphics[width=0.9\textwidth]{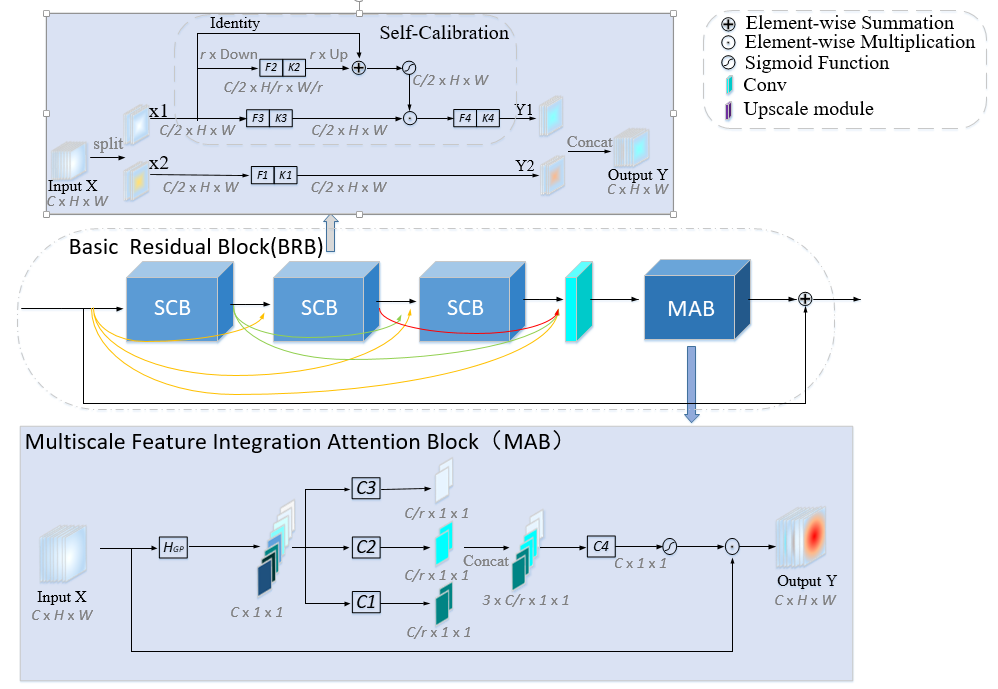}
    \caption{The proposed BRB and MAB for the OPPO\_CAMERA team. The top of the figure shows the basic convolution structure of the proposed network with the dense connection. The middle of the figure shows the basic residual block. The bottom of the figure presents the channel attention mechanism of the network.}
    \label{sc_block}
\end{figure}

\noindent
The OPPO\_CAMERA team proposed Self-Calibrated Attention Neural Network for Real-World Super Resolution. As shown in Fig.\ref{total_structure}, the proposed model is constituted of four integral components, \emph{i.e.}, feature extraction, residual in residual deep feature extraction, upsampling and reconstruction. It employs the same residual structure and dense connections to DRLN~\cite{anwar2019densely}. A longer skip connection is also added to connect the input to the output with a trainable parameter, which can greatly reduces the difficulty of optimization and thus, the network would pay more attention to the learning of the high frequency parts in images.
As shown in Fig.\ref{sc_block}, three Basic Residual Block (BRB) forms a Large Residual Block (LRB) with dense connection. Self-Calibration convolution (SCC)~\cite{liu2020improving}, shown at top of Fig.\ref{sc_block}, is adopted as a basic unit in order to expand receptive field. Unlike conventional convolution, SCC enables each point in space to have interactive information from nearby regions and channels. Dense connections are established between the Self-Calibration convolution block (SCCB), each densely connected residual block has three SCCB. To incorporate channel information efficiently, an attention block with multi-scale feature integration is added in every basic residual block as DRLN~\cite{anwar2019densely}.
For the network optimization, $L_1$ loss function was introduced as pixel-wise loss. In order to improve the fidelity, SSIM and MS-SSIM loss were also used as structure loss. With pixel loss and structure loss, the total loss is formulated as follows,
\begin{center}
$\mathcal{L}_{total} = \mathcal{L}^{{L}_1} + 0.2 \cdot \mathcal{L}^{MS-SSIM} + 0.2 \cdot \mathcal{L}^{SSIM}$
\end{center}

For the training, the proposed method splits the training data randomly into two parts, \emph{i.e.}, training set and validation set, with the ratio of 18500:500. Considering its significant improvement in the Real World SR task, CutBlur~\cite{yoo2020rethinking} is applied to augment training images. Self-ensemble and Parameter-fusion strategy would obviously improve the fidelity index(PSNR and SSIM), and meanwhile, less noise in result images. The strategy of self-ensembles (x8) was used as explained in RCAN~\cite{zhang2018image}, and all the corresponding parameters of last 3 models are fused to derive a fused model $G_{fused}$, as described in \cite{shang2020perceptual}. Experiments are conducted with Tesla V100 GPU.

\vspace{1em}
\noindent
\textbf{AiAiR}

\begin{figure}[t]
\centering
\includegraphics[width=4.2in]{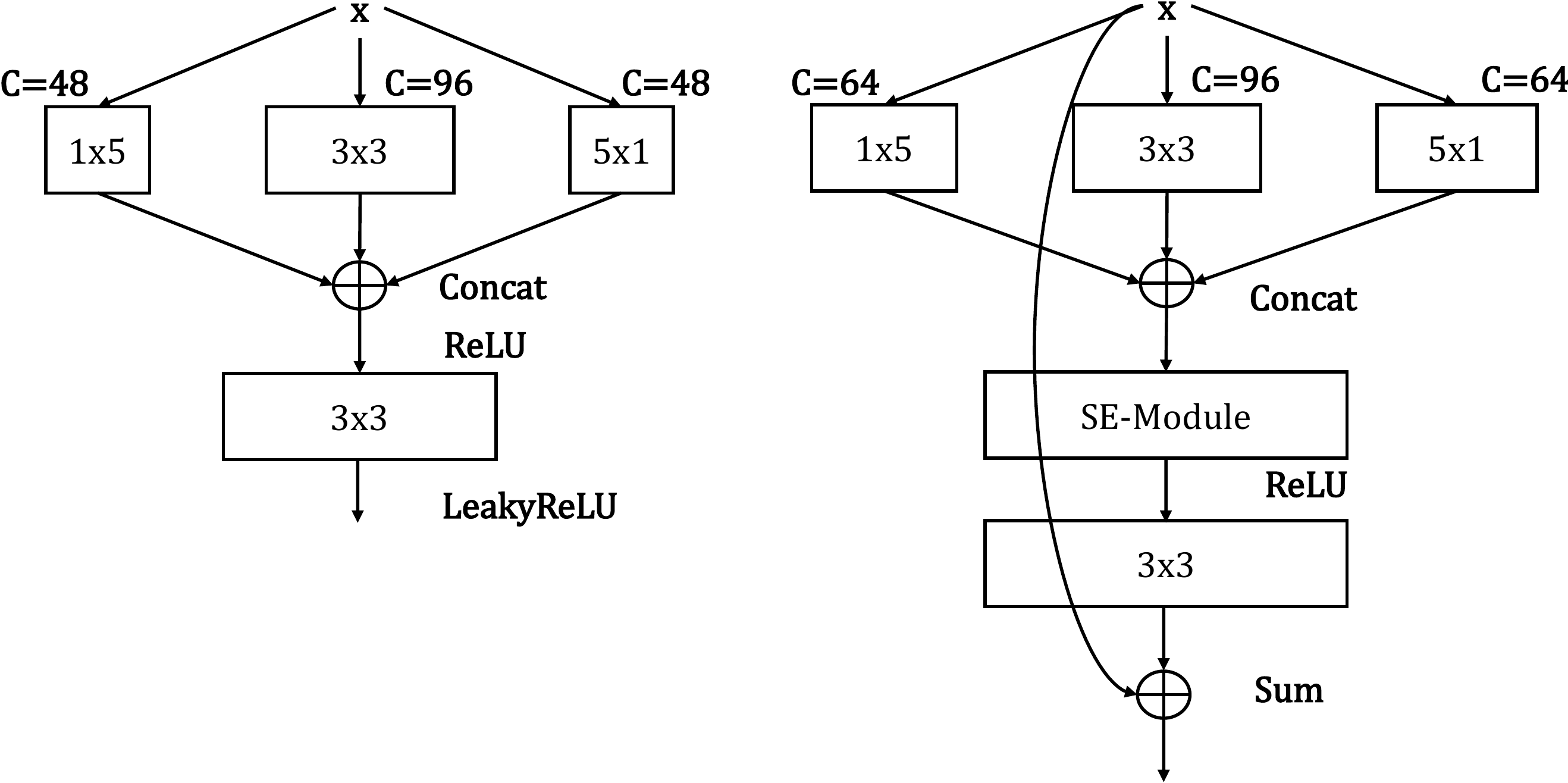}
\caption{OADDet and Deep-OADDet for the AiAiR team.}
\label{core_module}
\end{figure}
\begin{figure}[t]
\centering
\includegraphics[width=4.4in]{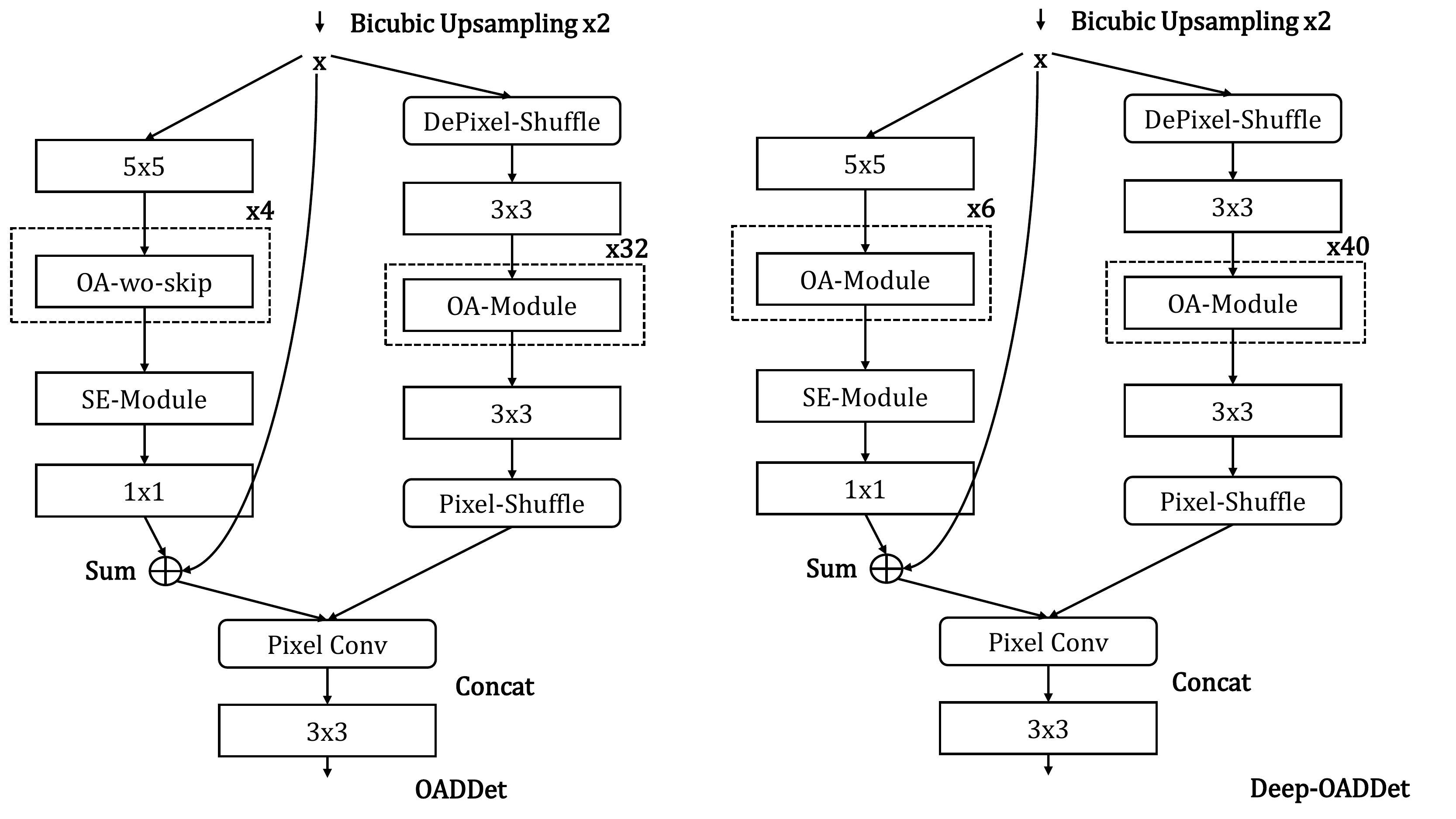}
\caption{Overall architectures of OADDet and Deep-OADDet for the AiAiR team.}
\label{network}
\end{figure}

\noindent
The AiAiR team proposes that orientation-aware convolutions meet dual path enhancement network (OADDet). Their method consists of four basic models (model ensemble): OADDet, Deep-OADDet, original EDSR \cite{Lim_2017_CVPR_Workshops} and original DRLN \cite{anwar2019densely}. The core modules of OADDet, illustrated in Figure \ref{core_module}, are borrowed from DDet \cite{shi2020ddet}, Inception \cite{inception} and OANet \cite{OANet} with minor improvements, such as less attention modules, removing skip connections and replacing ReLU with LeakyReLU. Overall architectures are similar to DDet \cite{shi2020ddet}. It is found that redundant attention modules will damage the performance and slow down the training process. Therefore, attention modules are only applied to the last few blocks of the backbone network and the last layer of the shallow network. Similar to RealSR \cite{cai2019toward}, PixelConv is also utilized, which contributes to $\sim 0.15$dB improvement on the validation set.
\begin{itemize}
\item The training process generally consists of four stages on three different datasets. The total training time is about 2000 GPU hours on V100.
\item OADDet models are trained from scratch and download DIV2K pre-trained EDSR/DRLN from official links. 
\item DIV2K dataset is used to pre-train our OADDet models and use manually washed AIM2020 datasets to fine-tune all models (further details in \href{https://github.com/HolmesShuan/AIM2020-RealSR}{GitHub README}).
\item Four models are trained using three different strategies:

1) For OADDet: Pre-training on DIV2K (300 epochs) then fine-tuning on original AIM2020 x2 dataset (600 epochs) and AIM2020 washed x2 dataset (100 epochs).

2) For Deep-OADDet: Pre-training on DIV2K (30 epochs) then fine-tuning on AIM2020 washed x2+x3 dataset (350 epochs), AIM2020 washed x2 dataset (350 epochs) and AIM2020 washed x2 dataset (100 epochs).

2) For EDSR/DRLN: Using DIV2K well-trained models then fine-tuning on washed AIM2020 x2 dataset (1000 epochs).

\item Self-ensemble ($\times8$), model-ensemble (four models) and proposed ``crop-ensemble'' are conducted (further details in \href{https://github.com/HolmesShuan/AIM2020-RealSR/blob/master/README.md#31-reproduce-x2-test-dataset-results}{GitHub README Reproduce x2 test dataset results}).
\item OADDet enjoys a more stable and faster training process than OANet, which introduces too many attention modules at the early stage of the networks. DDet proposes to use dynamic PixelConv with kernelsize=5,7,9; however, it is proved that kernelsize=3,5,7 works better during training and testing time.
\end{itemize}


\vspace{1em}
\noindent
\textbf{TeamInception}

\begin{figure}[!t]
\begin{center}
\begin{tabular}[t]{c} \hspace{-2mm}
\includegraphics[width=\textwidth]{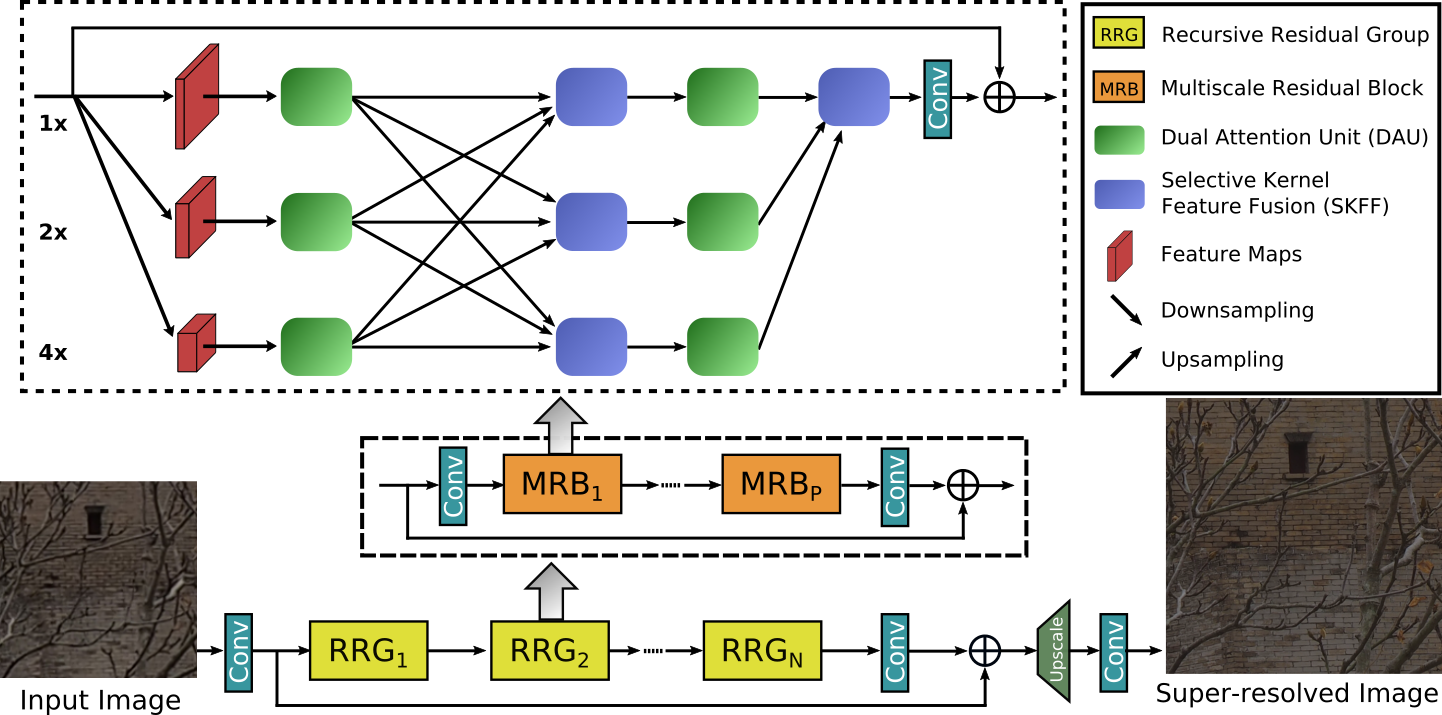}
\end{tabular}
\end{center}
\vspace*{-4mm}
\caption{\small Framework of the network MIRNet (recently introduced in \cite{Zamir2020MIRNet}) for the TeamInception team.}
\label{fig:framework}
\vspace{-1em}
\end{figure}

\noindent
The TeamInception team proposes learning Enriched Features for Real Image Restoration and Enhancement. MIRNet, recently introduced in \cite{Zamir2020MIRNet}, is utilized with the collective goals of maintaining spatially-precise high-resolution representations through the entire network and receiving strong contextual information from the low-resolution representations. In Fig.~\ref{fig:framework}. MIRNet\footnote{The code is publicly available at \url{https://github.com/swz30/MIRNet}} has a multi-scale residual block (MRB) containing several key elements:  \textbf{(a)} parallel multi-resolution convolution streams for extracting (fine-to-coarse) semantically-richer and (coarse-to-fine) spatially-precise feature representations, \textbf{(b)} information exchange across multi-resolution streams, \textbf{(c)} attention-based aggregation of features arriving from multiple streams, and \textbf{(d)} dual-attention units to capture contextual information in both spatial and channel dimensions.

The MRB consists of multiple (three in this work) fully-convolutional streams connected in parallel.
It allows information exchange across parallel streams in order to consolidate the high-resolution features with the help of low-resolution features, and vice versa. Each component of MRB is described as follows.

\begin{figure}[t]
\begin{center}
\scalebox{0.99}{
\begin{tabular}[t]{c} \hspace{-2mm}
\includegraphics[width=\textwidth]{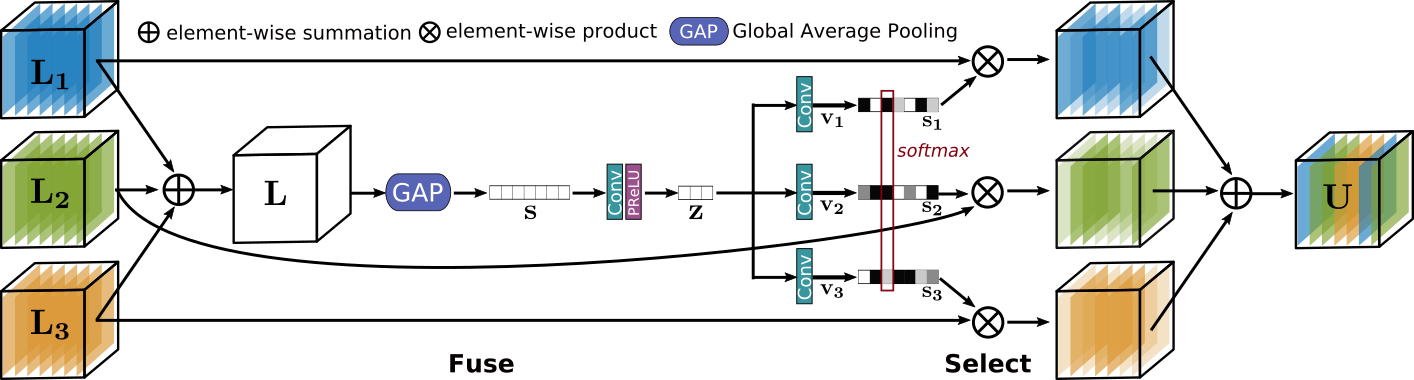}
\end{tabular}}
\end{center}
\vspace*{-6mm}
\caption{\small Schematic for selective kernel feature fusion (SKFF) for the TeamInception team. It operates on features from multiple convolutional.}
\label{fig:skff}
\vspace*{-0em}
\end{figure}

\textbf{Selective kernel feature fusion (SKFF).}
The SKFF module performs dynamic adjustment of receptive fields via two operations --{\emph{Fuse} and \emph{Select}, as illustrated in Fig.~\ref{fig:skff}}.
The \emph{fuse} operator generates global feature descriptors by combining the information from multi-resolution streams. The \emph{select} operator uses these descriptors to recalibrate the feature maps (of different streams) followed by their aggregation. Details of both operators for the three-stream case are elaborated as follows.
%
%
\textbf{(1) Fuse:} SKFF receives inputs from three parallel convolution streams carrying different scales of information.
We first combine these multi-scale features using an element-wise sum as: $\mathbf{L = L_1 + L_2 + L_3}$.
We then apply global average pooling (GAP) across the spatial dimension of $\mathbf{L} \in \mathbb{R}^{H\times W \times C}$ to compute channel-wise statistics $\mathbf{s} \in \mathbb{R}^{1\times 1 \times C}$.
Next, a channel-downscaling convolution layer is used to generate a compact feature representation $\mathbf{z} \in \mathbb{R}^{1\times 1 \times r}$, where $r=\frac{C}{8}$ for our experiments.
Finally, the feature vector $\mathbf{z}$ passes through three parallel channel-upscaling convolution layers (one for each resolution stream) and provides us with three feature descriptors $\mathbf{v_1}, \mathbf{v_2}$ and $\mathbf{v_3}$, each with dimensions $1\times1\times C$.
\textbf{(2) Select:} this operator applies the softmax function to $\mathbf{v_1}, \mathbf{v_2}$ and $\mathbf{v_3}$, yielding attention activations $\mathbf{s_1}, \mathbf{s_2}$ and $\mathbf{s_3}$ that we use to adaptively recalibrate multi-scale feature maps $\mathbf{L_1}, \mathbf{L_2}$ and $\mathbf{L_3}$, respectively. The overall process of feature recalibration and aggregation is defined as: $\mathbf{U = s_1 \cdot L_1 + s_2\cdot L_2 + s_3 \cdot L_3}$.
Note that the SKFF uses $\sim6\times$ fewer parameters than aggregation with the concatenation but generates more favorable results.

\begin{figure}[t]
\begin{center}
\scalebox{0.99}{
\begin{tabular}[t]{c} \hspace{-2mm}
\includegraphics[width=\textwidth]{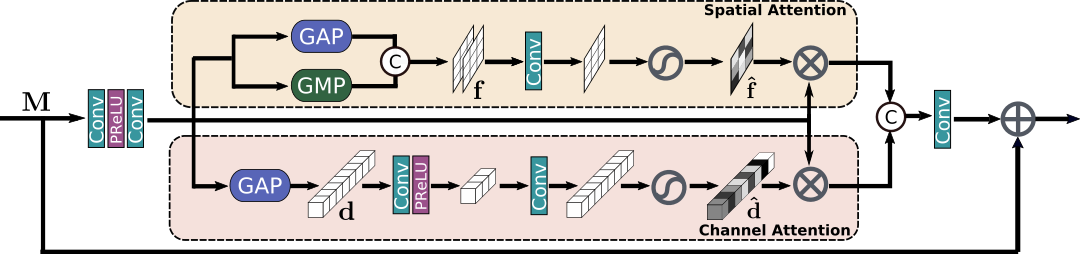}
\end{tabular}}
\end{center}
\vspace*{-6mm}
\caption{\small Dual attention unit incorporating spatial and channel attention mechanisms for the TeamInception team. }
\label{fig:dau}
\vspace{-0.6em}
\end{figure}

\vspace{0.4em} \noindent \textbf{Dual attention unit (DAU).}
While the SKFF block fuses information across multi-resolution branches, we also need a mechanism to share information within a feature tensor, both along the spatial and the channel dimensions.
The dual attention unit (DAU) is proposed to extract features in the convolutional streams. The schematic of DAU is shown in Fig.~\ref{fig:dau}.
The DAU suppresses less useful features and only allows more informative ones to pass further.
This feature recalibration is achieved by using channel attention~\cite{hu2018squeeze} and spatial attention~\cite{woo2018cbam} mechanisms.
\textbf{(1) Channel attention (CA)} branch exploits the inter-channel relationships of the convolutional feature maps by applying \emph{squeeze} and \emph{excitation} operations \cite{hu2018squeeze}. Given a feature map $\mathbf{M} \in \mathbb{R}^{H\times W \times C}$, the squeeze operation applies global average pooling across spatial dimensions to encode global context, thus yielding a feature descriptor $\mathbf{d} \in \mathbb{R}^{1\times 1 \times C}$.
The excitation operator passes $\mathbf{d}$ through two convolutional layers followed by the sigmoid gating and generates activations $\mathbf{\hat{d}} \in \mathbb{R}^{1\times 1 \times C}$.
Finally, the output of CA branch is obtained by rescaling $\mathbf{M}$ with the activations $\mathbf{\hat{d}}$.
\textbf{(2) Spatial attention (SA)} branch is designed to exploit the inter-spatial dependencies of convolutional features. The goal of SA is to generate a spatial attention map and use it to recalibrate the incoming features $\mathbf{M}$.
To generate the spatial attention map, the SA branch first independently applies global average pooling and max pooling operations on features $\mathbf{M}$ along the channel dimensions and concatenates the outputs to form a feature map $\mathbf{f} \in \mathbb{R}^{H\times W \times 2}$. The map $\mathbf{f}$ is passed through a convolution and sigmoid activation to obtain the spatial attention map $\mathbf{\hat{f}} \in \mathbb{R}^{H\times W \times 1}$, which is used to rescale $\mathbf{M}$.

\vspace{0.5em}
For training, $L_1$, multi-scale SSIM and VGG loss functions are considered in the model, defined as follows
\begin{align}
\mathcal{L}_{f} = \alpha \mathcal{L}_{1}(\hat{\mathbf{y}},\mathbf{y}) + \beta \mathcal{L}_{\text{MS-SSIM}}(\hat{\mathbf{y}},\mathbf{y}) + \gamma \mathcal{L}_{\text{VGG}}(\hat{\mathbf{y}},\mathbf{y})
\label{Eq:loss function}
\end{align}
$\mathcal{L}_{\text{VGG}}$ uses the features of \emph{conv2} layer after ReLU in the pre-trained VGG-16 network.
Three RRGs are utilized, each of which contains $2$ MRBs. MRB consists of $3$ parallel streams with channel dimensions of $64, 128, 256$ at resolutions $1, \frac{1}{2}, \frac{1}{4}$, respectively. Each stream has $2$ DAUs.
Patches with the size of $128\times128$  are cropped. Horizontal and vertical flips are employed for data augmentation. The model is trained from scratch with the Adam optimizer ($\beta_1 = 0.9$, and $\beta_2=0.999$) for $7\times10^5$ iterations. The initial learning rate is $2\times10^{-4}$ and the batch size is $16$. The cosine annealing strategy is employed to steadily decrease the learning rate from the initial value to $10^{-6}$ during training.


At inference time, the self-ensemble strategy [2] is employed. For each test image, a set of following 8 images are created: original, flipped, rotated $90^{\circ}$, rotated $180^{\circ}$, rotated $270^{\circ}$, $90^{\circ} \&$ flipped, $180^{\circ} \&$ flipped, and $270^{\circ} \&$ flipped. Next, these transformed images are passed through our model and obtain super-resolved outputs. Then we undo the transformations and perform averaging to obtain the final image. To fuse results, three different variants of the proposed networks are trained with different loss functions (Eq.~\ref{Eq:loss function}): \textbf{(1)} only the first term, \textbf{(2)} the first two terms (i.e., $\alpha \mathcal{L}_{1} +\beta \mathcal{L}_{\text{MS-SSIM}}$), and \textbf{(3)} all the terms.
For the variant 2, $\alpha = 0.16$ and $\beta = 0.84$; for the variant 3, $\alpha = 0.01$ and $\beta = 0.84$, $\gamma = 0.15$.

Given an image, the generated self-ensembled results with each of these three networks are averaged to obtain the final image.
Results with self-ensemble strategy and fusion are reported in Table~\ref{Table:self-ensemble}. With 4 Tesla-V100 GPUs, it takes $\sim$3 days to train the network. The time required to process a test image of size $3780\times5780$ is 2 seconds (single method), 30 seconds (self-ensemble) and 87 seconds (fusion).

\begin{table*}[!t]
\centering
\setlength{\tabcolsep}{8pt}
\caption{Results of validation set for the scale factor $\times4$ for the TeamInception team. Comparison of using single method (SM), self-ensemble (SE) and Fusion (F) on validation set. }
\begin{tabular}{l c c c c}
\toprule
  & $\mathcal{L}_{1}$ & $\mathcal{L}_{1} + \mathcal{L}_{\text{MS-SSIM}}$ & $\mathcal{L}_{1} + \mathcal{L}_{\text{MS-SSIM}} + \mathcal{L}_{\text{VGG}}  $ & PSNR\\
\midrule
\multicolumn{1}{l}{SM} & $\surd$ &  &  & 29.72 \\
\multicolumn{1}{l}{SM} &  & $\surd$ &  & 29.83  \\
\multicolumn{1}{l}{SM} &  &  & $\surd$ & 29.89 \\
\multicolumn{1}{l}{SM + F} & $\surd$ & $\surd$ & $\surd$ & 30.08 \\
\multicolumn{1}{l}{SE + F} & $\surd$ & $\surd$ & $\surd$ & 30.25 \\
\bottomrule
\end{tabular}
\label{Table:self-ensemble}
\end{table*}

\vspace{1em}
\noindent
\textbf{Noah\_TerminalVision}

\noindent
The Noah\_TerminalVision team proposed Super Resolution with weakly-paired data using an Adaptive Robust Loss. The network is based on RRDBNet with 23 Residual in Residual Denseblocks. Only training pairs with a high PSNR score were used for training. To further alleviate the bad effect of miss-alignment of training data, the adaptive robust loss function proposed by Jon Barron was used. For track 3, it additionally used a spatial attention module and an efficient channel attention module. The spatial attention module is borrowed from EDVR~\cite{wang2019edvr} and the efficient attention module is borrowed from ECA-Net~\cite{wang2020eca}. Considering that the training data are not perfectly aligned, Adaptive Robust Loss Function \cite{barron2019general} for super resolution tasks is utilized to solve the weakly-paired training problem.
The self-ensemble strategy is to run inference on the combination of the 90/180/270-degree rotated images of the original/flipped input and then to average the results.


Only training pairs with a high PSNR score (29) were used for training. The learning rate is 2e-4, the patch size of inputs is $80\times80$ and the batchsize is 4. CosineAnnealingLR\_Restart learning rate scheme is employed and the restart period is 250,000 steps. For each input, due to GPU memory constraint, images are tested patch-wisely. The crop window is of size 120$\times$120, and a stride of 110$\times$110 was used to collect patches.


\vspace{1em}
\noindent
\textbf{DeepBlueAI}

\noindent
The DeepBlueAI team proposed a solution based on RCAN~\cite{RCAN}, which was implemented with PyTorch. In each RG, the RCAB number is 20, G=10 and C=128 in the RIR structure. The model is trained from scratch, which costs about 4 days with 4$\times$32G Tesla V100 GPU. For training, all the training images are augmented by random horizontal flips and 90 rotations. In each training batch, LR color patches with the size of 64$\times$64 are extracted as inputs. The initial leaning rate is set to $2.0\times 10^{-4}$ and learning rate of each parameter group use a cosine annealing schedule with total $1.0\times 10^{5}$ iterations and without restart. For testing, each low resolution image is flipped and rotated to generate seven augmented inputs; with the trained RCAN model, the corresponding super-resolved images are generated. An inverse transform is applied to those output images to get the original geometry. The transformed outputs are averaged all together to yield the self-ensemble result.

\vspace{1em}
\noindent
\textbf{ALONG}

\begin{figure*}[t]
\centering
\includegraphics[width=0.9\linewidth]{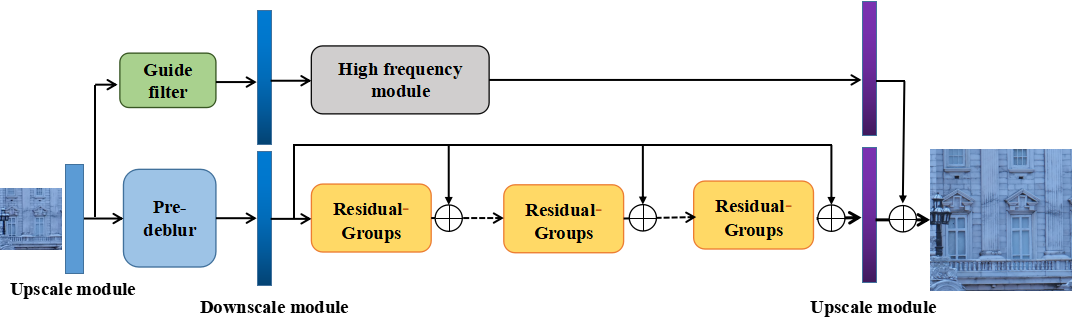}
\caption{RCAN for the Real Image Super-Resolution (RCANv2) for the ALONG team.}
\label{fig:fig1}
\end{figure*}

\noindent
The ALONG team proposed Dual Path Network with high frequency guided for real-world image Super-Resolution. The proposed method follows the main structure of RCAN~\cite{RCAN} and utilizes the guild filter to decompose the detail layer and to restore high-frequency details. As illustrated in Figure~\ref{fig:fig1}, a lot of share-source skip connections in the original feature extraction path with channel attention. Due to share-source skip connections, the abundant low-frequency information can be bypassed and facilitate to train deeper network.
Compared with the previous simulated datasets, the image degradation process for real SR is much more complicated. Low-resolution images lose more high-frequency information and look blurry. Inspired by other image deblurring tasks~\cite{wang2019edvr,zhou2019davanet,zhou2019spatio}, a pre-deblur module is used before the residual groups to pre-process blurry inputs and improve super-resolution accuracy. Specifically, the input image is first down-sampled with strided convolution layers; then the upsampling layer at the end will resize the features back to the original input resolution.
The proposed dual path network restores fine details by decomposing the input image and focusing on the detail layers. An additional branch focuses on the high-frequency reconstruction. The input LR image is decomposed into the detail layer using the guided filter, an edge-preserving low-pass filter~\cite{he2012guided}. Then a high-frequency module is adopted on the detail layer, so the output result can focus on restoring high-frequency details.

Besides, a variety of data augmentation strategies are combined to achieve competitive results in different tracks, including Cutout~\cite{devries2017improved}, CutMix~\cite{yun2019cutmix}, Mixup~\cite{zhang2017mixup}, CutMixup, RGB permutation, Blend. In addition, inspired by~\cite{yoo2020rethinking}, CutBlur, unlike Cutout, can utilize the entire image information while it enjoys the regularization effect due to the varied samples of random HR ratios and locations. The experimental results also show that a reasonable combination of data enhancement can improve the model performance without additional computation cost in the test phase.
The model is trained with 8 2080Ti, 11G memory each GPU. Pseudo ensemble is also employed. The inputs are flipped/rotated and the HR results are aligned and averaged for enhanced prediction.

\vspace{1em}
\noindent
\textbf{LISA-ULB}

\begin{figure}[h]
	\begin{center}
		\includegraphics[width=0.8\linewidth]{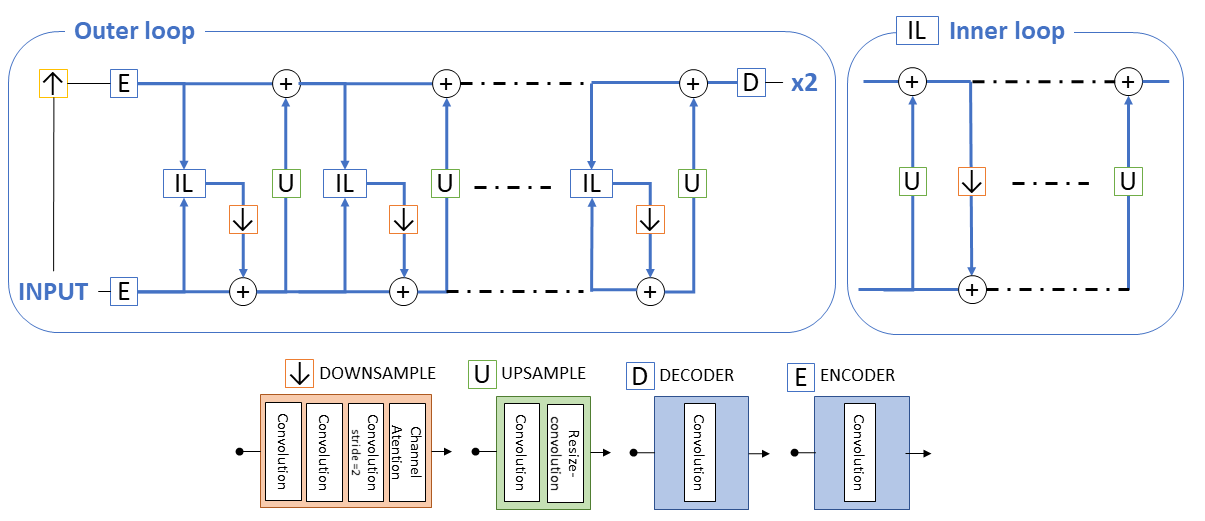}
	\end{center}
	\caption{The architecture of the proposed network by the LISA-ULB team.}
	\label{fig:LISA-ULB}
\end{figure}

\noindent
The LISA-ULB team proposed VCycles BackProjection networks generation two (VCBPv2), which utilized an iterative error correcting feedback mechanism to guide the reconstruction of the final SR output. As shown in Figure~\ref{fig:LISA-ULB}, the proposed network is composed of an outer loop of 10 cycles and an inner loop of 3 cycles. The input of the proposed VCBPv2 is the LR image and the upsampled counterpart. The upsample and downsample modules iteratively transform features between high- and low-resolution space as residual for error correction. The decoder in the end reconstructs the corrected feature to SR image.

The model is trained using AdamW optimizer with learning rate of $1\times 10^{-4}$ and halved at every 400 epochs, then the training is followed by SGDM optimizer. Equally weighted $\ell_{1}$ and SSIM loss is adopted for training.

\vspace{1em}
\noindent
\textbf{lyl}

\begin{figure}[h]
    \begin{center}
        \includegraphics[width=0.4\linewidth]{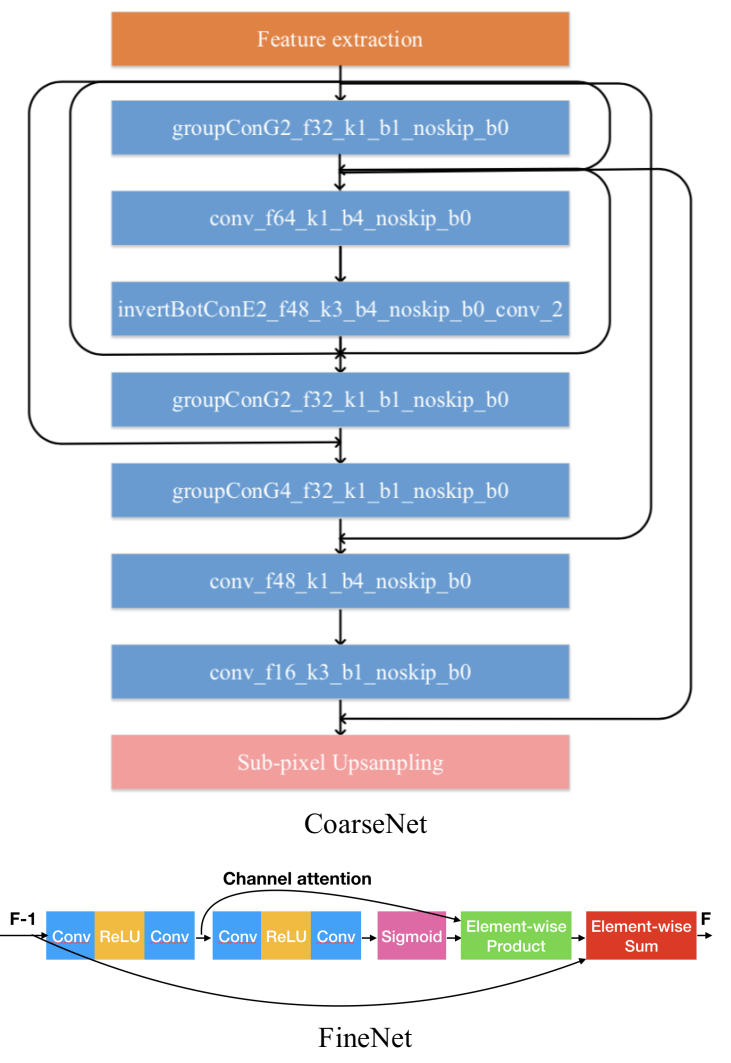}
    \end{center}
\caption{The architecture of the proposed network by the lyl team.}
\label{fig:lyl}
\end{figure}

\noindent
The lyl team proposed a coarse to fine network for progressive super-resolution. As shown in Fig\ref{fig:lyl}, based on the Laplacian pyramid framework, the proposed model takes an LR image as input and progressively predicts residual images at $S_{1},S_{2}...S_{n}$ levels. $S$ is the scale factor, $S= S_{1}\times S_{2}...\times S_{n}$, where $n= log_{2}^{S}$.

$\ell_{1}$ was adopted to optimize the proposed network. Each level of the proposed CFN was supervised by different scales of HR images.

\vspace{1em}
\noindent
\textbf{GDUT-SL}

\noindent
The GDUT-SL team used the RRDBNet of ESRGAN\cite{ESRGAN} to perform super-resolution. Typical RRDB block has 3 Dense blocks, which including 5 Conv layers with Leaky-ReLU and remove BN layers. The RRDB number was set to 23. Two UpConv layer is used for upsampling. Different from ESRGAN, the GDUT-SL team replaced the activation function with ReLU to obtain better PSNR results.

Residual scaling and smaller initialization were adopted to facilitate training a deep architecture. In training phase, the mini-batch size was set to 16, with image size of 96$\times$96. 20 promising models were selected for model-ensemble.

\vspace{1em}
\noindent
\textbf{MCML-Yonsei}

\begin{figure}[h]
\centering
        \includegraphics[width=1\linewidth]{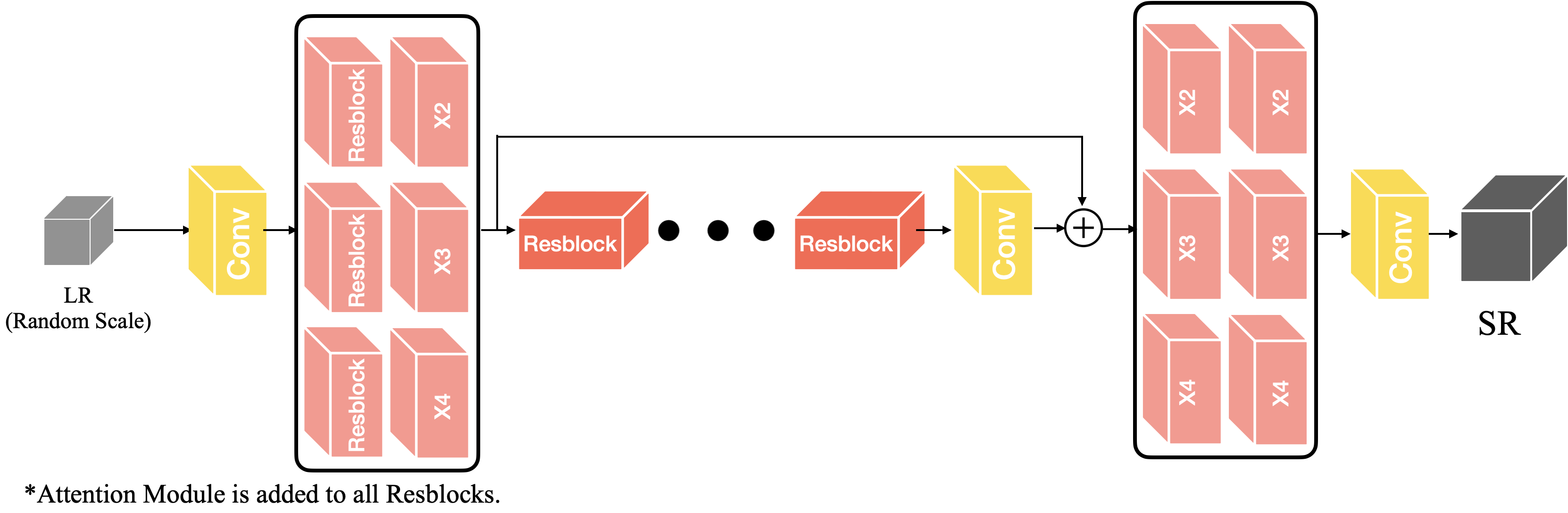}
\caption{Overview of the network for the MCML-Yonsei team.}
\label{fig:MCML_model}
\end{figure}

\begin{figure}[h]
\centering
        \includegraphics[width=1\linewidth]{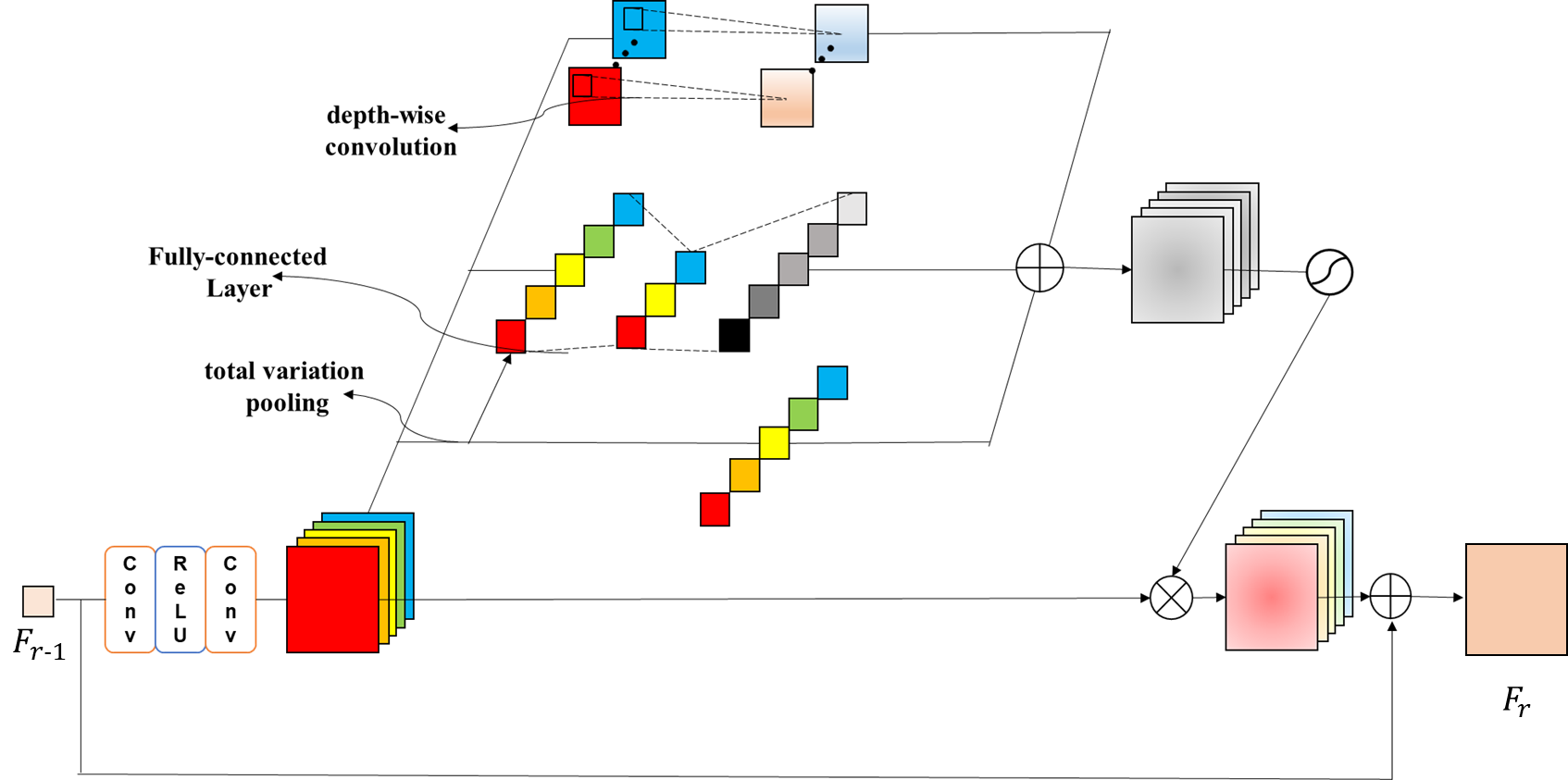}
\caption{Resblock with Attention Module for the MCML-Yonsei team.}
\label{fig:MCML_rblock}
\end{figure}

\noindent
As shown in Fig.\ref{fig:MCML_model}, the MCML-Yonsei team proposed an attention based multi-scale deep residual network based on MDSR~\cite{EDSR}, which shares most of the parameters across different scales. In order to utilize various features in each real image adaptively, the MCML-Yonsei team added an attention module in the existing Resblock. As shown in Fig.\ref{fig:MCML_rblock}, the attention module is based on MAMNet~\cite{MAMNet} where the global variance pooling was replaced with total variation pooling.

They initialized all parameters except the attention module with the pre-trained MDSR, which was optimized for bicubic downsampling based training data. The mini-batch size was set to 16 and the patch size was set to 48. They subtracted the mean of each R, G, B channel of the train set for data normalization. The learning rate was initially set to $1e-4$, and it decayed at the 15k steps. The total training step was 20k.

\vspace{1em}
\noindent
\textbf{kailos}

\begin{figure}[h]
    \begin{center}
        \includegraphics[width=0.6\linewidth]{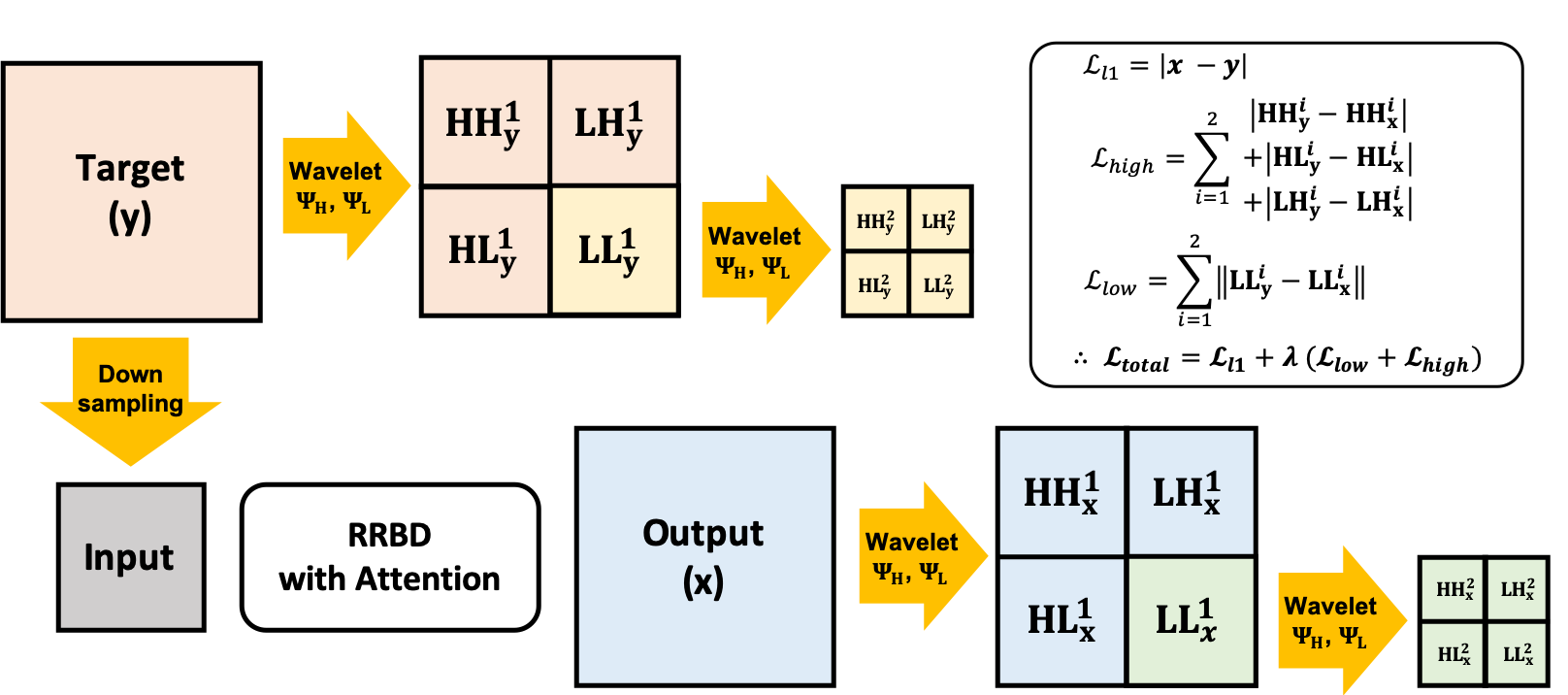}
    \end{center}
\label{fig:kailos}
\caption{Overview of the proposed method of for the kailos team.}
\end{figure}

\noindent
The kailos team proposed RRBD Network with Attention mechanism using Wavelet loss for Single Image Super-Resolution. The loss function consisted of conventional $L_1$ loss $\mathcal{L}_{L_1}$ and novel wavelet loss $\mathcal{L}_{wavelet}$. The conventional $L_1$ loss $\mathcal{L}_{L_1}$ is given as
$\mathcal{L}_{L_1} = \sum{\mid x - y \mid_{1}}$, where $x$ is reconstructed image and $y$ is ground truth image.

A wavelet transform can separate the signal features along the low and high frequency components. Most of the energy distribution in the signal, such as global structure and color distribution, is concentrated in the low frequency components. On the other hand, the high frequency components include signal patterns and image textures. Since both frequency components have different characteristics, a different loss function must be applied to each component. Therefore, the proposed novel wavelet loss $\mathcal{L}_{wavelet}$ is the sum of $L_1$ loss for high frequency components and $L_2$ loss for low frequency components given as $\mathcal{L}_{high} = \sum_{i=1}^{N}{\mid\Psi_{H}^i(x) - \Psi_{H}^i(y) \mid_{1}}$, $\mathcal{L}_{low} = \sum_{i=1}^{N}{\parallel \Psi_{L}^i(x) - \Psi_{L}^i(y) \parallel_{2}^{2}}$, and $\mathcal{L}_{wavelet} = \mathcal{L}_{low} + \mathcal{L}_{high}$, where $N$ denotes the stage of wavelet transform and $\Psi_{H}$ and $\Psi_{L}$ are high and low frequency decomposition filters, respectively.

In the experiment, $N$ is 2 and Haar wavelet filters are used as wavelet decomposition filters. Therefore, a total loss is defined by $\mathcal{L}_{total} = \mathcal{L}_{L_1} + \lambda~\mathcal{L}_{wavelet}$, where $\lambda$ denotes the regularization parameter and $\lambda = 1$ was used in the proposed method.  Fig.\ref{fig:kailos} shows an overview of the proposed method. Adam optimizer was used in training process, and the size of image patch was the quarter size of training data.

\vspace{1em}
\noindent
\textbf{qwq}

\begin{figure}[h]
\centering
        \includegraphics[width=1\linewidth]{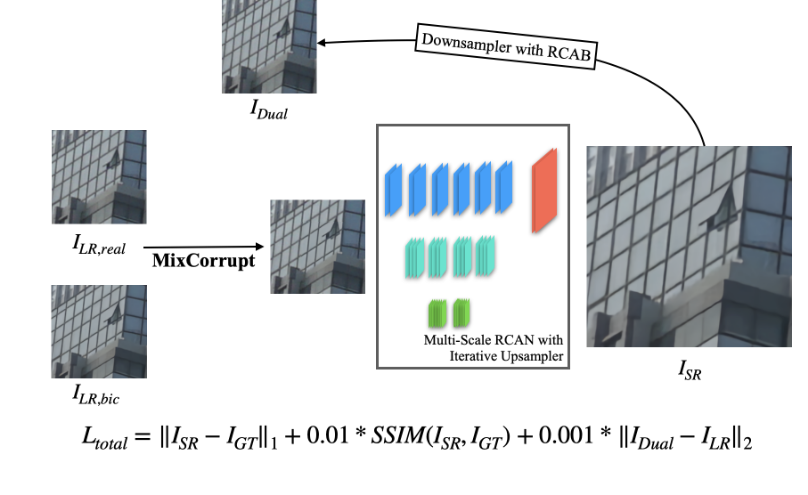}
\caption{The total learning diagram of for the qwq team. In upsample network, they used features from 0.25$\times$, 1$\times$, 2$\times$ and 4$\times$(HR) five scales.}
\label{fig:kailos}
\end{figure}

\noindent
The qwq team proposed a Multi-Scale Network based on RCAN\cite{RCAN}. As shown in Fig[1], the multi-scale mechanism was integrated into the base block of RCAN in order to enlarge the receptive field. Dual Loss was adopted for training. MixCorrupt augmentation was conducted, for it allowed the network to learn from robust SR results from different degradations, which is specially designed for the real-world scenario.

\vspace{1em}
\noindent
\textbf{RRDN\_IITKGP}

\noindent
The RRDN\_IITKGP used a GAN based Residual in Residual Dense Network~\cite{ESRGAN}, where the model is pre-trained on other dataset and evaluated on the challenge dataset.

%
%
%
%
%

\vspace{1em}
\noindent
\textbf{SR-IM}

\begin{figure}[h]
\centering
        \includegraphics[width=0.8\linewidth]{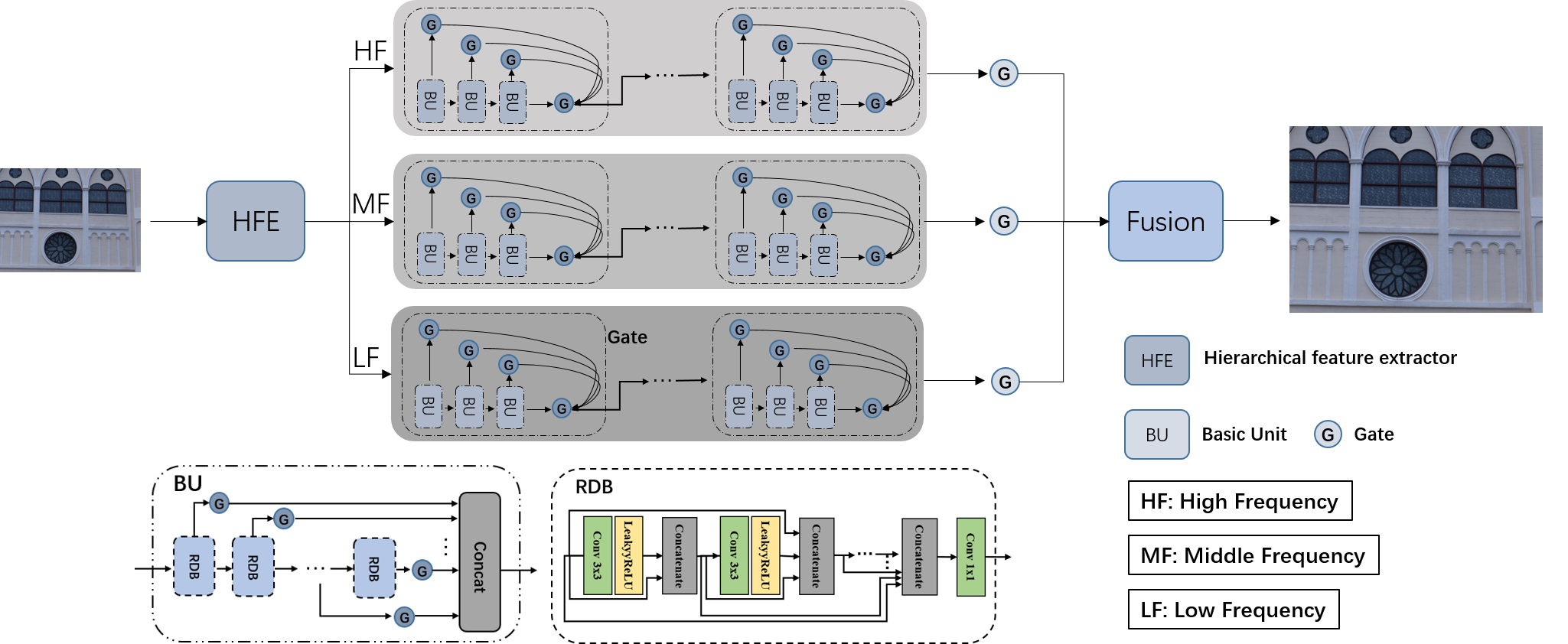}
\caption{Structure of Frequency-aware Network (FAN) for the SR-IM team. There are three branches, representing the high frequency, middle frequency and low frequency components. The gate attention is used to adaptively select the required frequency components.}
\label{fig:SR-IM}
\end{figure}

\noindent
The SR-IM team proposed frequency-aware network, as shown in Fig.\ref{fig:SR-IM}. A hierarchical feature extractor (HFE) is utilized to extract the high representation, middle representation and low representation. The basic unit of the body consists of residual dense block and channel attention module. Finally, the three branches are fused into one super-resolved image by the gate and fusion module.

The mini-batch size was set to 8 and the patch size was set to 160 during training. They used Adam optimizer with an initial learning rate of 0.0001. The learning rate decayed by a factor of 0.5 every 30 epochs. The entire training time is about 48 hours.

\vspace{1em}
\noindent
\textbf{JNSR}

\noindent
The JNSR team utilized EDSR~\cite{EDSR} and DRLN~\cite{anwar2019densely} to perform model ensemble. The EDSR and DRLN were trained on AIM2020 dataset, the best models were chosen for model ensemble.

\vspace{1em}
\noindent
\textbf{SR\_DL}

\begin{figure}[h]
    \begin{center}
        \includegraphics[width=0.8\linewidth]{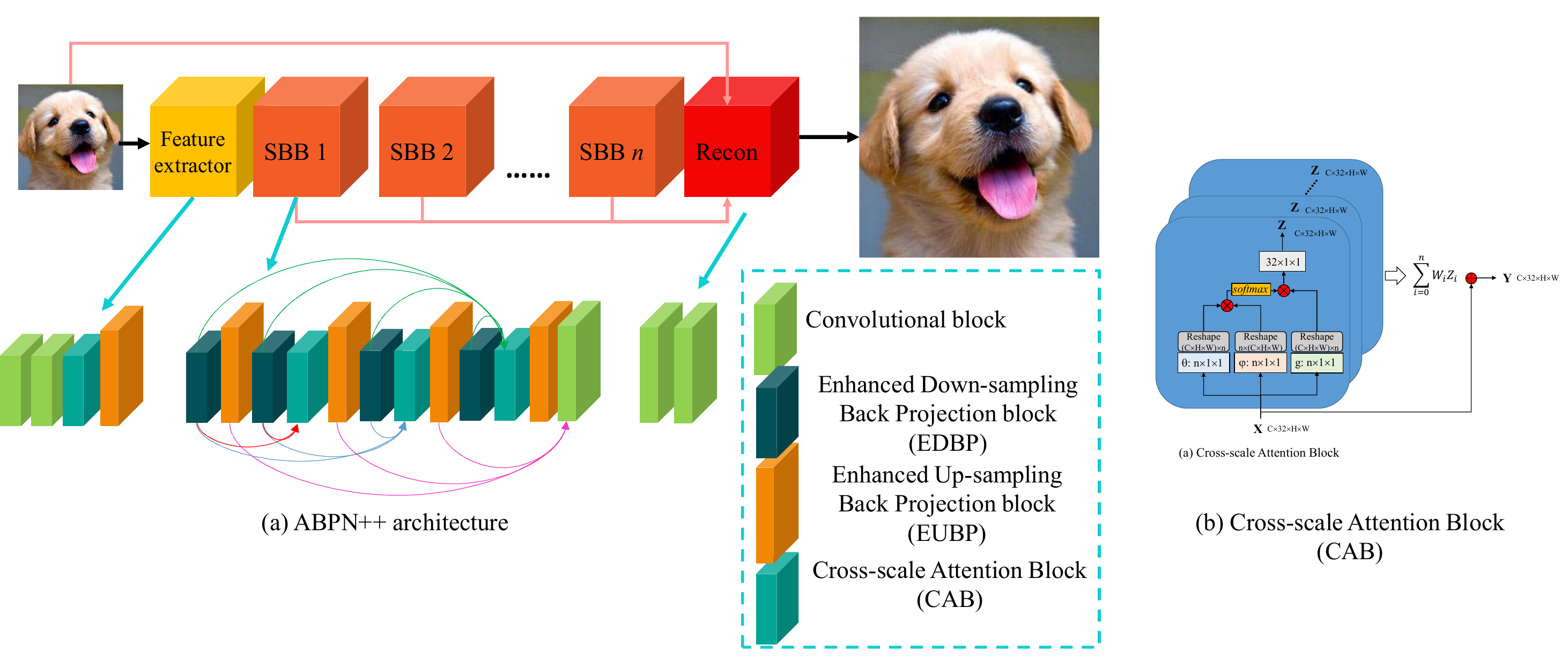}
    \end{center}
\caption{ (a): ABPN++: Attention based Back Projection Network for image super-resolution. (b): the proposed Cross-scale Attention Block by the SR\_DL team.}
\label{fig:SRDL}
\end{figure}

\noindent
The SR\_DL team proposed attention back projection network (ABPN++), as shown in Fig.\ref{fig:SRDL}. The proposed ABPN++ network first conducts feature extraction to expand the feature space of the input LR image. Then the densely connected enhanced down- and up-sampling back projection blocks perform up- and down-sampling the feature maps. The Cross-scale Attention Block (CAB) takes the outputs from down-sampling back projection blocks to compute the cross-correlation for feature fusion. Finally, the Refined Back Projection Block works as a final refinement that estimates the feature residuals between input LR and predicted LR images for update. The complete network includes 10 down- and up-sampling back projection block, 2 feature extraction blocks and 1 refined back projection block. Each back projection block is made of 5 convolutional layers. The kernel number is 32 for all convolution and deconvolution layers. For down- and up-sampling convolution layer, the kernel size is 6, stride is 4 and padding is 1.

The mini-batch size was set to 16 and the LR patch size was set to 48 during training. The learning rate is fixed to 1e-4 for all layers for $2\times 10^5$ iterations in total as the first stage. Then the batch size increases to 32 for $1\times 10^5$ iterations as fine-tuning.

\vspace{-0.5em}

\noindent
\textbf{Webbzhou}

\noindent
The Webbzhou team fine-tuned the pre-trained RRDB~\cite{ESRGAN} on the challenge dataset.

\vspace{0.6em}
\noindent
\textbf{MoonCloud}

\noindent
The MoonCloud team utilized RCAN~\cite{RCAN} for the challenge. Totally 6 models were used for model ensemble. Three of them were trained on challenge dataset with scale of 4. The other three were trained on the challenge dataset with scale of 3, which were fine-tuned on the dataset with scale of 4 after. The final outputs were obtained by averaging the outputs of these six models.

\vspace{0.5em}
\noindent
\textbf{SrDance}

\noindent
The SrDance team utilized RRDB~\cite{ESRGAN}. A new training strategy was adopted for model optimization. The model was firstly pre-trained on DIV2K dataset. Then they trained their model by randomly picking one image in dataset and randomly crop a few $40\times 40$ patches, which is alike stochastic gradient descent. Second, when model stepped, they trained on 10 pics, one $40\times 40$ patch from each picture and fed to the model.

\vspace{0.5em}
\noindent
\textbf{MLP\_SR}

\begin{figure}[h]
    \begin{center}
        \includegraphics[width=0.4\linewidth]{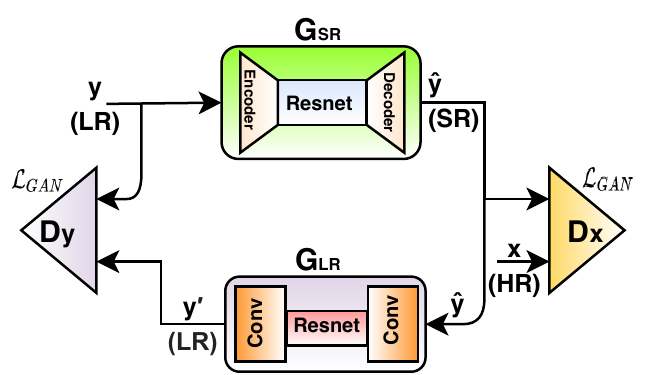}
    \end{center}
\caption{Illustration of the structure of SR approach setup proposed by the MLP\_SR team.}
\label{fig:MLPSR}
\end{figure}

\noindent
The MLP\_SR team proposed Deep Cyclic Generative Adversarial Residual Convolutional Networks for Real Image Super-Resolution, as shown in Fig.\ref{fig:MLPSR}. The SR generator~\cite{Umer2020DeepGA} network $G_{SR}$ was trained in a GAN framework by using the LR ($\mathbf{y}$) images with their corresponding HR images with pixel-wise supervision in the clean HR target domain ($\mathbf{x}$), while maintaining the cyclic consistency between the LR and HR domain.

\vspace{1em}
\noindent
\textbf{congxiaofeng}

\noindent
The congxiaofeng team proposed RDB-P SRNet, which contains several residual-dense blocks with pixel shuffle for upsampling. The network was inspired by RDN~\cite{RDN}.

\vspace{1em}
\noindent
\textbf{debut\_kele}

\noindent
The debut\_kele team proposed Enhanced Deep Residual Networks for real image super-resolution.

\section*{Acknowledgements}

We thank the AIM 2020 sponsors: Huawei, MediaTek, Google, NVIDIA, Qualcomm and Computer Vision Lab (CVL) ETH Zurich.
This work was partially supported from National Key Research and Development Project, Fundamental Research Funds for the Central Universities under Grant No.19lgpy228, China Postdoctoral Science Foundation (2020M672968).

\vspace{1em}
\appendix
\section*{A. Teams and affiliations}\label{app:A}
\textbf{AIM2020 team}

\noindent
\emph{Title:} AIM 2020 Real Image Super-Resolution Challenge

\noindent
\emph{Members:}

\noindent
Pengxu Wei$^1$ (\email{weipx3@mail.sysu.edu.cn}),

\noindent
Hannan Lu$^2$ (\email{hannanlu@hit.edu.cn}),

\noindent
Radu Timofte$^3$ (\email{radu.timofte@vision.ee.ethz.ch}),

\noindent
Liang Lin$^1$ (\email{linliang@ieee.org}),

\noindent
Wangmeng Zuo$^2$ (\email{cswmzuo@gmail.com})

\noindent
\emph{Affiliations:}

\noindent
$^1$ Sun Yat-sen University

\noindent
$^2$ Harbin Institute of Technology University

\noindent
$^3$ Computer Vision Lab, ETH Zurich, Switzerland

~\\
\noindent
\textbf{Baidu}

\noindent
\emph{Title:} Real Image Super Resolution via Heterogeneous Model Ensemble using GP-NAS

\noindent
\emph{Members:} Zhihong Pan$^1$ (\email{zhihongpan@baidu.com}),
Baopu Li$^1$
Teng Xi$^2$,
Yanwen Fan$^2$,
Gang Zhang$^2$,
Jingtuo Liu$^2$,
Junyu Han$^2$,
Errui Ding$^2$

\noindent
\emph{Affiliation:}

\noindent
$^1$ Baidu Research (USA)

\noindent
$^2$ Department of Computer Vision Technology (VIS), Baidu Incorportation

~\\
\textbf{CETC-CSKT}

\noindent
\emph{Title:} Adaptive dense connection super resolution reconstruction

\noindent
\emph{Members:} Tangxin Xie (\email{xxh96@outlook.com}),
Yi Shen,
Jialiang Zhang,
Yu Jia,
Liang Cao,
Yan Zou

\noindent
\emph{Affiliation:}
%
China Electronic Technology Cyber Security Co., Ltd.

~\\
\textbf{OPPO\_CAMERA}

\noindent
\emph{Title:} Self-Calibrated Attention Neural Network for Real-World Super Resolution

\noindent
\emph{Members:} Kaihua Cheng (\email{chengkaihua@oppo.com}),
Chenhuan Wu

\noindent
\emph{Affiliation:}
%
Guangdong OPPO Mobile Telecommunications Corp., Ltd

~\\
\textbf{ALONG}

\noindent
\emph{Title:} Dual Path Network with High Frequency Guided for Real World Image Super-Resolution

\noindent
\emph{Members:} Yue Lin (\email{gzlinyue@corp.netease.com}),
Cen Liu,
Yunbo Peng

\noindent
\emph{Affiliation:}
%
NetEase Games AI Lab

~\\
\textbf{Noah\_TerminalVision}

\noindent
\emph{Title:} Super Resolution with weakly-paired data using an Adaptive Robust Loss

\noindent
\emph{Members:} Xueyi Zou (\email{zouxueyi@huawei.com}),

\noindent
\emph{Affiliation:}
Noah's Ark Lab, Huawei

~\\
\textbf{DeepBlueAI}

\noindent
\emph{Title:} A solution based on RCAN

\noindent
\emph{Members:} Zhipeng Luo, Yuehan Yao (\email{yaoyh@deepblueai.com}), Zhenyu Xu

\noindent
\emph{Affiliation:} DeepBlue Technology (Shanghai) Co., Ltd

\noindent

~\\
\textbf{TeamInception}

\noindent
\emph{Title:} Learning Enriched Features for Real Image Restoration and Enhancement

\noindent
\emph{Members:} Syed Waqas Zamir (\email{waqas.zamir@inceptioniai.org}),
Aditya Arora,
Salman Khan,
Munawar Hayat,
Fahad Shahbaz Khan

\noindent
\emph{Affiliation:}
%
Inception Institute of Artificial Intelligence (IIAI)
\vspace{2em}

~\\
\textbf{MCML-Yonsei}

\noindent
\emph{Title:} Multi-scale Dynamic Residual Network Using Total Variation for Real Image Super-Resolution

\noindent
\emph{Members:} Keon-Hee Ahn (\email{khahn196@gmail.com}),
Jun-Hyuk Kim,
Jun-Ho Choi,
Jong-Seok Lee

\noindent
\emph{Affiliation:}
%
Yonsei University

~\\
\textbf{lyl}

\noindent
\emph{Title:} Coarse to Fine Pyramid Networks for Progressive image super-resolution

\noindent
\emph{Members:} Tongtong Zhao (\email{daitoutiere@gmail.com}),
Shanshan Zhao

\noindent
\emph{Affiliation:}
%
Dalian Maritime Univerity

~\\
\textbf{kailos}

\noindent
\emph{Title:} RRDB Network with Attention mechanism using Wavelet loss for Single Image Super-Resolution

\noindent
\emph{Members:} Yoseob Han$^1$ (\email{yoseobhan@lanl.gov}),
Byung-Hoon Kim$^2$,
JaeHyun Baek$^3$

\noindent
\emph{Affiliation:}

\noindent
$^1$ Loa Alamos National Laboratory (LANL)

\noindent
$^2$ Korea Advanced Institute of Science and Technology (KAIST)

\noindent
$^3$ Amazon Web Services (AWS)

~\\
\textbf{qwq}

\noindent
\emph{Title:} Dual Learning for SR using Multi-Scale Network

\noindent
\emph{Members:} Haoning Wu, Dejia Xu
%
\emph{Affiliation:} Peking University

\noindent

~\\
\textbf{AiAiR}

\noindent
\emph{Title:} OADDet: Orientation-aware Convolutions Meet Dual Path Enhancement Network

\noindent
\emph{Members:} Bo Zhou$^1$ (\email{1826356001@qq.com}),

\noindent
Haodong Yu$^2$ (\email{haodong.yu@outlook.com})

\noindent
\emph{Affiliation:}

\noindent
$^1$ Jiangnan University

\noindent
$^2$ Karlsruher Institut fuer Technologie

~\\
\textbf{JNSR}

\noindent
\emph{Title:} Dual Path Enhancement Network

\noindent
\emph{Members:} Bo Zhou (\email{jeasonzhou1@gmail.com})

\noindent
\emph{Affiliation:}
%
Jiangnan University

~\\
\textbf{SrDance}

\noindent
\emph{Title:} Training Strategy Optimization

\noindent
\emph{Members:} Wei Guan (\email{missanswer@163.com}),
Xiaobo Li,
Chen Ye

\noindent
\emph{Affiliation:}
%
Tongji University

~\\
\textbf{GDUT-SL}

\noindent
\emph{Title:} Ensemble of RRDB for Image Restoration

\noindent
\emph{Members:} Hao Li (\email{2111903004@mail2.gdut.edu.cn}),
Haoyu Zhong,
Yukai Shi,
Zhijing Yang,
Xiaojun Yang

\noindent
\emph{Affiliation:}
%
Guangdong University of Technology

~\\
\textbf{MoonCloud}

\noindent
\emph{Title:} Mixed Residual Channel Attention

\noindent
\emph{Members:} Haoyu Zhong (\email{hy0421@outlook.com}),
Yukai Shi,
Xiaojun Yang,
Zhijing Yang,

\noindent
\emph{Affiliation:}
%
Guangdong University of Technology,

~\\
\textbf{SR-IM}

\noindent
\emph{Title:} FAN: Frequency-aware network for image super-resolution

\noindent
\emph{Members:} Xin Li (\email{lixin666@mail.ustc.edu.cn}),
Xin Jin,
Yaojun Wu,
Yingxue Pang,
Sen Liu

\noindent
\emph{Affiliation:}
%
University of Science and Technology of China

~\\
\textbf{SR\_DL}

\noindent
\emph{Title:} ABPN++: Attention based Back Projection Network for image super-resolution

\noindent
\emph{Members:} Zhi-Song Liu$^1$ (\email{zhisong.liu@connect.polyu.hk}),
Li-Wen Wang$^2$,
Chu-Tak Li$^2$,
Marie-Paule Cani$^1$,
Wan-Chi Siu$^2$

\noindent
\emph{Affiliation:}

\noindent
$^1$ LIX - Computer science laboratory at the Ecole polytechnique [Palaiseau]

\noindent
$^2$ Center of Multimedia Signal Processing, The Hong Kong Polytechnic University

~\\
\textbf{Webbzhou}

\noindent
\emph{Title:} RRDB for Real World Super-Resolution

\noindent
\emph{Members:}Yuanbo Zhou (\email{webbozhou@gmail.com}),

\noindent
\emph{Affiliation:}
%
Fuzhou University, Fujian Province, China

~\\
\textbf{MLP SR}

\noindent
\emph{Title:} Deep Cyclic Generative Adversarial Residual Convolutional Networks for Real Image Super-Resolution

\noindent
\emph{Members:} Rao Muhammad Umer (\email{engr.raoumer943@gmail.com}),
Christian Micheloni

\noindent
\emph{Affiliation:}
%
University Of Udine, Italy

~\\
\textbf{congxiaofeng}

\noindent
\emph{Title:} RDB-P SRNet: Residual-dense block with pixel shuffle

\noindent
\emph{Members:} Xiaofeng Cong (\email{1752808219@qq.com})

\noindent
\emph{Affiliation:} (Not provided)

\noindent

~\\
\textbf{RRDN\_IITKGP}

\noindent
\emph{Title:}  A GAN based Residual in Residual Dense Network

\noindent
\emph{Members:} Rajat Gupta (\email{rajatgba2021@email.iimcal.ac.in})

\noindent
\emph{Affiliation:}
%
Indian Institute of Technology

%
%
%

~\\
\textbf{debut\_kele}

\noindent
\emph{Title:} Self-supervised Learning for Pretext Training

\noindent
\emph{Members:} Kele Xu (\email{kelele.xu@gmail.com}),
Hengxing Cai,
Yuzhong Liu

\noindent
\emph{Affiliation:}
%
National University of Defense Technology

~\\
\textbf{Team-24}

\noindent
\emph{Title:} VCBPv2 - VCycles Backprojection Upscaling Network

\noindent
\emph{Members:} Feras Almasri (\email{Feras.Almasri@ulb.ac.be}),
Thomas Vandamme,
Olivier Debeir

\noindent
\emph{Affiliation:} Universi\'e Libre de Bruxelles, LISA department

\noindent

%

\clearpage
%
%
\bibliographystyle{splncs04}
\bibliography{egbib}
\end{document}